\documentclass[runningheads]{llncs}
\usepackage{graphicx}
\graphicspath{ {./images/} }
\usepackage{comment}
\usepackage{amsmath,amssymb} % define this before the line numbering.
\usepackage{color}

\usepackage[dvipsnames]{xcolor}
\usepackage[normalem]{ulem}

\usepackage{url}
\usepackage{verbatim}

\usepackage{caption}
\usepackage{subcaption}
\usepackage{capt-of}
\usepackage{makecell}

\newcommand{\Figref}[1]{Fig.~\ref{#1}}
\newcommand{\Tblref}[1]{Table~\ref{#1}}
\newcommand{\Secref}[1]{Sec.~\ref{#1}}
\usepackage[width=122mm,left=12mm,paperwidth=146mm,height=193mm,top=12mm,paperheight=217mm]{geometry}

\begin{document}
% \renewcommand\thelinenumber{\color[rgb]{0.2,0.5,0.8}\normalfont\sffamily\scriptsize\arabic{linenumber}\color[rgb]{0,0,0}}
% \renewcommand\makeLineNumber {\hss\thelinenumber\ \hspace{6mm} \rlap{\hskip\textwidth\ \hspace{6.5mm}\thelinenumber}}
% \linenumbers
\pagestyle{headings}
\mainmatter

\title{Depth Completion with RGB Prior} % Replace with your title

% INITIAL SUBMISSION 
%\begin{comment}
% \titlerunning{ECCV-20 submission ID \ECCVSubNumber} 
% \authorrunning{ECCV-20 submission ID \ECCVSubNumber} 
\author{Yuri Feldman \textsuperscript{1}, Yoel Shapiro \textsuperscript{2}, Dotan Di Castro \textsuperscript{2}}
\institute{
    \textsuperscript{1} Technion - Israel Institute of Technology \newline
    \url{yurif@cs.technion.ac.il} \newline
    \leavevmode\newline
    \textsuperscript{2} Bosch Center for Artificial Intelligence \newline
    \url{{Shapiro.Yoel, Dotan.DiCastro}@bosch.com}
}
%\end{comment}
%******************

% CAMERA READY SUBMISSION
\begin{comment}
\titlerunning{Depth Completion with RGB Prior}
% If the paper title is too long for the running head, you can set
% an abbreviated paper title here
%
\author{First Author\inst{1}\orcidID{0000-1111-2222-3333} \and
Second Author\inst{2,3}\orcidID{1111-2222-3333-4444} \and
Third Author\inst{3}\orcidID{2222--3333-4444-5555}}
%
\authorrunning{F. Author et al.}
% First names are abbreviated in the running head.
% If there are more than two authors, 'et al.' is used.
%
\institute{Princeton University, Princeton NJ 08544, USA \and
Springer Heidelberg, Tiergartenstr. 17, 69121 Heidelberg, Germany
\email{lncs@springer.com}\\
\url{http://www.springer.com/gp/computer-science/lncs} \and
ABC Institute, Rupert-Karls-University Heidelberg, Heidelberg, Germany\\
\email{\{abc,lncs\}@uni-heidelberg.de}}
\end{comment}
%******************
\maketitle

%\ys{at least 70 and at most 150 words}
\begin{abstract}
Depth cameras are a prominent perception system for robotics, especially when operating in natural unstructured environments. Industrial applications, however, typically involve reflective objects under harsh lighting conditions, a challenging scenario for depth cameras, as it induces numerous reflections and deflections, leading to loss of robustness and deteriorated accuracy. Here, we developed a deep model to correct the depth channel in RGBD images, aiming to restore the depth information to the required accuracy. To train the model, we created a novel industrial dataset that we now present to the public. The data was collected with low-end depth cameras and the ground truth depth was generated by multi-view fusion. 
\end{abstract}

\section{Introduction}\label{sec:introduction}

% Motivation
Robotic manipulation has been an active research field for a few decades and is gaining renewed traction with novel learning-based methods. Industrial robotics has especially benefited from Deep Reinforcement Learning (DRL; \cite{schoettler2019deep}) and deep computer vision. A typical setup for industrial systems is a robotic arm served by a perception system that drives a visual signal into the robot's high-level controller. Industrial applications, and manufacturing in particular, require precise localization and pose estimation for successful grasping of the manipulated object. This enhanced perception requirement is often catered by depth cameras \cite{ten2018using} which yield RGBD images, i.e. RGB and an additional depth map.

% Describing the problem
Here, we investigate the usage of a 3D camera in an industrial use-case. By using a relatively inexpensive 3D camera\footnote{We use Intel RealSense 415 3D camera. } worth around \$150, we collected over 3,500 samples of 45 different industrial objects. Industrial scenarios are predominantly reflective since they are composed of smooth metallic objects, including parts, tools, and jigs. Under these conditions, many low-end 3D cameras provide poor depth data with high noise and many pixels with invalid values. These issues arise from limitations of the Infra-Red (IR) beam, which is used in Time-of-Flight (ToF) depth cameras to scan the field of view or to produce patterns for point-correlations in structured-light depth cameras. Reflections between scene components can confuse the sensors and create artifacts. For example, a reflective table top may appear to have dents, mirroring the objects laid upon it. Attentive work station design can minimize such reflections, but they cannot be entirely eliminated and are considered difficult to handle. A more pervasive problem is encountered on corners and edges of reflective objects, where the IR beams often deflect away from the sensor's field of view, leaving a "hole" in the depth channel. To alleviate the missing values issue, depth camera manufacturers usually provide post-processing algorithms, including smoothing filters and interpolations. However, the accuracy of these traditional methods tends to fall short of robotic manufacturing requirements. The system we used calculates depth using stereo cameras, enhanced by a static-pattern IR projector. High-end system can achieve better accuracy and robustness, for example by projecting a series of carefully crafted patterns at the cost of a longer acquisition time. Moreover, the cost of high-end systems can reach two orders of magnitude greater. Taken together, high end depth cameras are not an acceptable solution for most industrial applications.

% Explain the alternative and its limitations
An alternative to depth-cameras is multi-view perception, using ten or more different view-points in space (and possibly time) \cite{scharstein2002taxonomy,lazaros2008review,tippetts2016review}. The most notable variants of multi-view methods are \textit{Simultaneous Localization and Mapping} (SLAM; \cite{mur2017orb,handa2014benchmark}) and \textit{Structure from Motion} (SfM; \cite{faugeras1988motion,huang2002motion}), which are quite popular for mobile robots. However, SLAM and SfM are inclined to intensive computations and are inherently time-consuming, raising doubts on their applicability to industrial robotics and manufacturing. In the opposite direction, previous studies on monocular depth estimation applied a deep model on a single view image to estimate depth. While monocular depth estimation has been successful in several domains \cite{godard2017unsupervised,jrgensen2019monocular,kuznietsov2017semi}, it has not yet reached sufficient accuracy for industrial applications.

A promising avenue is to employ depth cameras for acquisition, followed by post-processing by depth corrections algorithms that exploit the RGB information as a prior for the correct depth values. In the next section, we provide details on two notable works \cite{Zhang18cvpr,sajjan2019cleargrasp} that follow this direction. These studies break down the solution into two steps. The first step applies a deep model on the RGB information to estimate surface normals and additional tasks. The second step uses optimization techniques to find a depth surface that fulfills a mixed objective, combined from the depth information and the outputs of the first step.

% Our approach and solution
In this work, we use an end-to-end deep learning (DL; \cite{lecun2015deep}) approach where we train a network to complete the depth in the missing places. We applied an encoding-decoding architecture, with a ResNet \cite{he2016deep} or VGG  \cite{simonyan2014very} backbone, adding a U-net style skip connections \cite{ronneberger2015u} between the corresponding encoder and decoder layers. We collected industrial real-world training data with a static rig of multiple (4) depth cameras. The ground truth depth maps are generated using TSDF \cite{Curless96accgit}, which is an imperfect approximation of the true depth values. Nonetheless, our model is trained to improve the input depth to the level of the attainable ground truth estimation. In the results section (\Secref{sec:results}), we quantify the average error between the input and ground-truth and show that our model reduces it by 1-2 orders of magnitude. Overall, our contributions are two-fold: 
\begin{enumerate}
    \item We provide a depth completion dataset of several thousand samples for a few dozen industrial objects, primarily challenging and specular objects. The data consists of RGBD measurements from a cost-effective device, embodying the real-world issues which hinder industrial adoption of depth cameras. We also introduce the fused 3D meshes, a by-product of our ground-truth generation algorithm.
    \item We present a depth-correction model for the industrial manufacturing domain. By incorporating ideas from previous studies and adjacent domains, we show that our model achieves better accuracy than the closest competitor (see \Secref{sec:results}).
\end{enumerate}
In the next section (\Secref{sec:related_work}), we review related work and describe the research landscape. In \Secref{sec:methods}, we expand on the methods we applied and how we crafted them to attain the best result. In \Secref{sec:results}, we describe our results and provide a discussion. We conclude the work (\Secref{sec:conclusions}) by summarizing our contributions and proposing future work directions.

\section{Related Work}\label{sec:related_work}

In this section, we review previous work related to depth completion, all of which make use of the RGB information. We begin from monocular depth estimation from RGB, which clearly demonstrates how pertinent RGB is to depth. We continue to sparse-to-dense depth completion, which has a lot in common with our task and demonstrates how much is achievable with only a sparse sample of depth information. We conclude with dense depth completion, where the depth data is dense but has missing or erroneous regions.

% \ys{is this cited? seems highly relevant}
% Learning Common Representation from RGB and Depth Images

% \ys{consider adding refs}
% 3D Photography using Context-aware Layered Depth Inpainting
% "https://shihmengli.github.io/3D-Photo-Inpainting/?fbclid=IwAR2ydr8YC_00a9-ETDh9Jm5KqpP_mvncnjooDadueKgap6J-qY1Xe_gplwY"

%%%%%%%%%%%%%%%%%%%%%%%%%%%%%%%%%%%%%%%%%%%

\subsection{Single-Frame Monocular Depth Estimation from RGB}
\label{subsec:monocular}

An extensive body of work provided a multitude of solutions for monocular depth estimation from RGB images. The premise is that RGB contains all the information needed to estimate depth. This notion is supported by the observation that people can estimate 3D from RGB using prior knowledge on objects shapes and sizes, shading and obstruction cues, etc. Furthermore, RGB cameras are considered remarkably cost effective, ubiquitous, and resilient to phenomena that adversely affect active acquisition modalities such as depth-cameras.

% 1, 2, 3
In \cite{Liu15cvpr}, the authors employ Continuous Random Fields (CRF) and Convolutional Neural Network (CNN) in order to estimate depth. They over-segment the image into super pixels and predict a depth value for the super-pixel. Their CRF objective is closely related to image matting \cite{image_matting}. In \cite{Eigen14nips}, the authors used a pair of coarse-to-fine deep networks. The first network is a CNN with a Fully Convolutional (FC) head that predicts a coarse depth map. The input image and the intermediate coarse outputs are fed into a Fully Convolutional Network (FCN) that yields a refined depth map. In \cite{Laina16ic3dv}, the authors use an encoder-decoder architecture \cite{Hinton504} with a Residual Network (ResNet 50; \cite{he2016deep}) backbone. They introduce a novel up-convolution block, a hybrid of upsampling and convolution, which allows addressing several (4) scales concurrently at a given layer. Other multi-scale blocks can be seen in \cite{GoogLeNet,Res2Net_Gao_2019}.

%4
Several works \cite{Garg16eccv,zhou2017unsupervised,faugeras1988motion,huang2002motion,godard2017unsupervised} developed unsupervised training pipelines, exploiting either prior knowledge of camera motion in combination with multiple-view geometry or SfM techniques. Here, we applied multiple-geometry to generate an explicit ground-truth and train in a supervised manner. Some of these unsupervised-training works solved a subsidiary task, for example warping a left stereo image to match the right image. Multi Task Learning (MTL) has been used in top scoring works (e.g. \cite{Kendall18cvpr}), and it is considered to be beneficial by its own merit.

% In \cite{Garg16eccv}, an interesting approach was taken: the problem is still to reconstruct depth from RGB image. The problem is that not always one have abundant labeled data in order to train a network. Therefore, they train the network with known camera motion between two close frames. As a result, since the displacement is known, one can estimate the depth from the motion.

%5
% The work of \cite{zhou2017unsupervised} uses movies in order to estimate depth. Such techniques are a particular case of the more general technique of structure from motion \cite{faugeras1988motion,huang2002motion}. A middle approach for static images is \emph{stereo vision} as in \cite{godard2017unsupervised} where using disparity estimation, neural networks were trained to warp left images to match the right.

% \begin{comment}
% 	\item[Kendall18cvpr{Kendall18cvpr}] Multi-Task Learning Using Uncertainty to Weigh Losses for Scene Geometry and Semantics
% 	\item[Zhou17cvpr~\cite{Zhou17cvpr}] Unsupervised learning of depth and ego-motion from video
% \end{comment}

%%%%%%%%%%%%%%%%%%%%%%%%%%%%%%%%%%%%%%%%%%%

\subsection{Sparse-to-Dense Depth Completion}
\label{subsec:sparse-to-dense}

% \yf{most of these works tested on KITTI, except for Zhang18arxiv ?} \yf{Recent \emph{completion} works target mostly Kitti / lidar setting (i.e. very sparse depth, usually more or less randomly sampled (or a sparse grid). Exceptions are \cite{Zhang18arxiv,sajjan2019cleargrasp}}.  

A related task is predicting a dense depth image, given its corresponding RGB image and sparse samples of the target depth image, around 5\% of the pixels. This problem setting originates in driving applications, for sensor fusion of RGB cameras and LiDAR scanners. This setting is akin to a regression problem, where a piece wise continuous solution needs to satisfy the sampled depth points. Most of the sparse-to-dense works test on the outdoor driving KITTI dataset \cite{Uhrig17ic3dv} and a few test additionally on the indoor NYUv2 dataset \cite{Silberman:ECCV12}, after simulating the LiDAR sampling behavior.
A dominant architecture in this domain is Encoder-Decoder \cite{Ma18icra,Ma19icra,Yang19cvpr} and sometimes even a double Encoder-Decoder \cite{Giannone19cvpr_workshop,kuznietsov2017semi}. In \cite{Ma19icra}, the Encoder-Decoder was embellished with U-connections \cite{UNet}, where the authors introduced a smoothness loss term and self-supervised training techniques. 

% 3, 4, 5
Other groups employed Multi Task Learning (MTL), including posterior probability of the prediction \cite{Yang19cvpr} (which is analogues to confidence), semantic segmentation \cite{Giannone19cvpr_workshop}, and image alignment \cite{kuznietsov2017semi}. The later \cite{kuznietsov2017semi} applied supervised training for depth prediction and added the image-alignment loss in an unsupervised manner. In \cite{Giannone19cvpr_workshop}, five different tasks were provided: semantic segmentation from RGB, semantic segmentation from depth (D), semantic segmentation from RGBD, depth prediction from RGB (monocular), and sparse-to-dense depth completion.

\subsection{Dense Depth Completion}
\label{subsec:dense-depth-completion}

Depth cameras typically fail on reflective objects such as metals and mirrors, and deflective objects such as glass objects and clear plastics. Specular objects result in connected regions (blobs) of missing or erroneous depth values, while the remaining depth map retains its integrity. This comprehension is reflected in auxiliary tasks, shared across the projects described in this subsection. The first commonality is using RGB to partition the images; for example, in \cite{Eigen15iccv} the model predicts semantic segmentation and in \cite{Zhang18cvpr,sajjan2019cleargrasp}  occlusion boundaries. The model in \cite{sajjan2019cleargrasp} predicts additionally a binary mask for clear objects. The second commonality is using the RGB to produce an initial estimation for the depth, only with a surprising twist. Unlike monocular models, \cite{Eigen15iccv,Zhang18cvpr,sajjan2019cleargrasp} predict surface normals, i.e. the gradient of the depth which in some sense can be thought of as a "spatial residual". Both in \cite{Zhang18cvpr} and \cite{sajjan2019cleargrasp}, the input is RGBD, unlike \cite{Eigen15iccv} which only used RGB. They applied a two-stage solution: the first stage consists of two or three parallel encoder-decoders applied only to the RGB data. The outputs are passed on to the second stage along with the depth data. The second stage resorts to global optimization for predicting the depth map. The optimizer minimizes an objective that corresponds to the surface normals and the valid regions of the depth input. In \cite{sajjan2019cleargrasp}, the first stage provides an explicit mask for invalid regions, while in \cite{Zhang18cvpr} it seems to be implicitly embedded in the depth input since the corrupted regions have outlying values. Notably, \cite{sajjan2019cleargrasp} provides a novel dataset for transparent objects with high quality depth maps. The dataset consists of over 50k synthetic images (5 objects) and 500 real world images (4 objects) collected in a painstaking process of replacing transparent objects with opaque replicas. In light of this, it is compelling to incorporate surface normal prediction.

\section{Methods}\label{sec:methods}

% \begin{figure}
%     \centering
%     \includegraphics[width=\linewidth]{images/sample.blue.png}
%     \caption{Sample from our dataset. Left column: input rgb and depth. Right column: fused rgb and depth. }
%     \label{fig:sample}
% \end{figure}

\begin{figure}
    \centering
    % Background white
    \includegraphics[width=.2\linewidth]{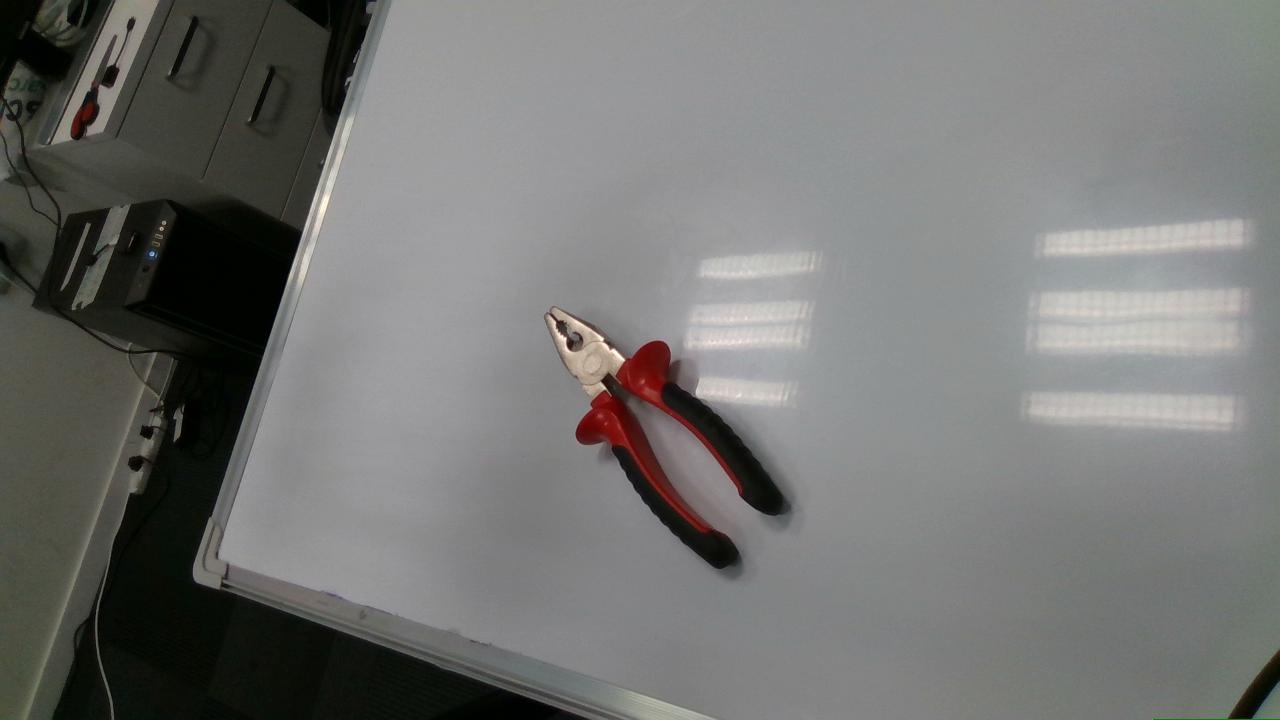}
    \includegraphics[width=.2\linewidth]{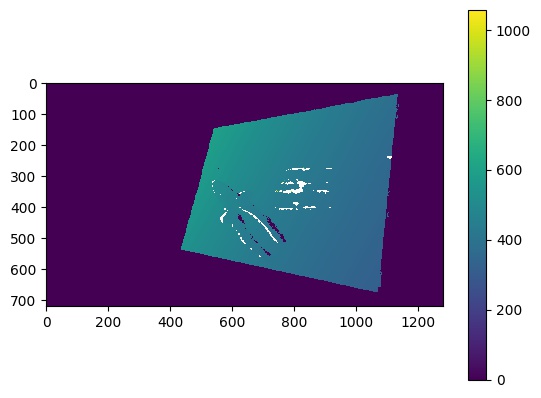}
    \includegraphics[width=.2\linewidth]{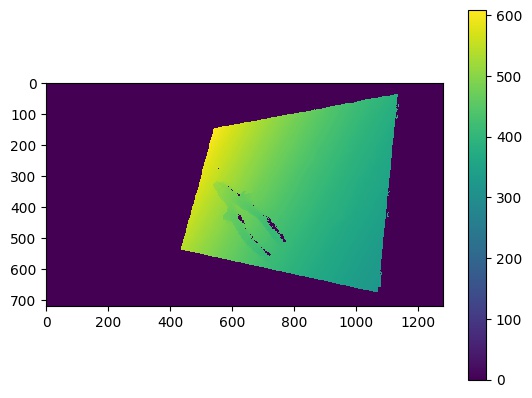}
    \includegraphics[width=.2\linewidth]{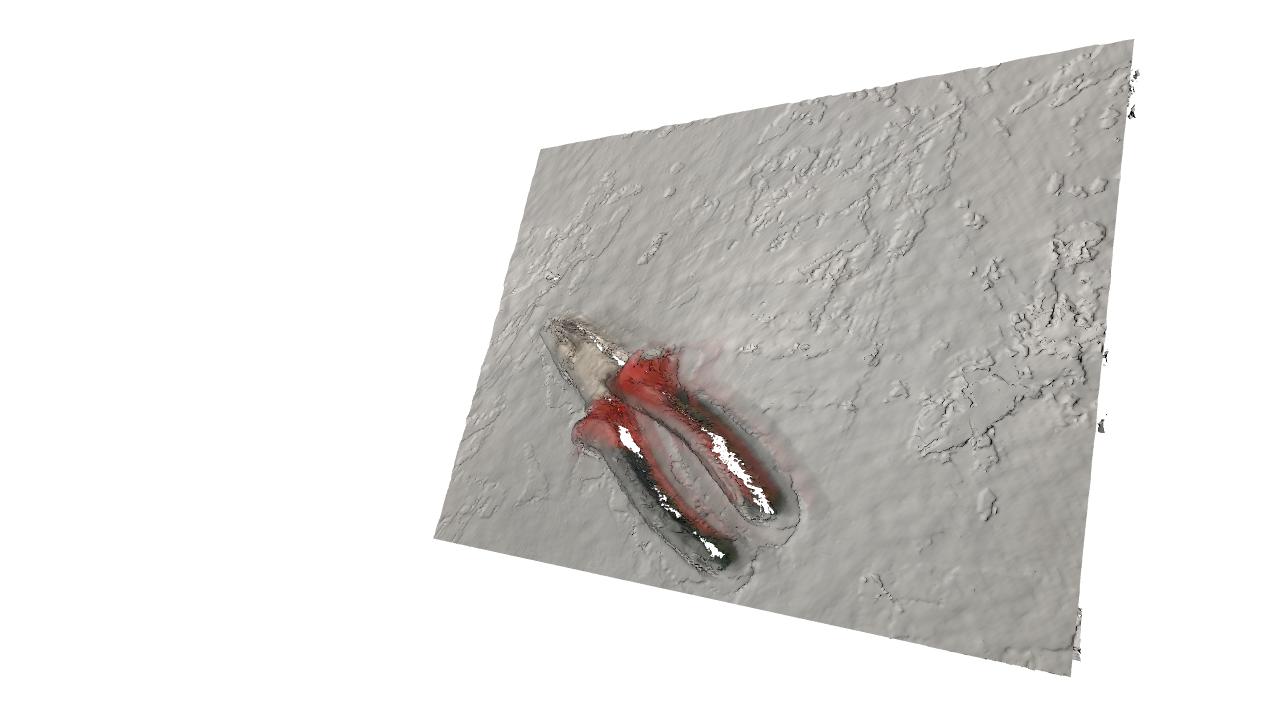}
    
    \includegraphics[width=.2\linewidth]{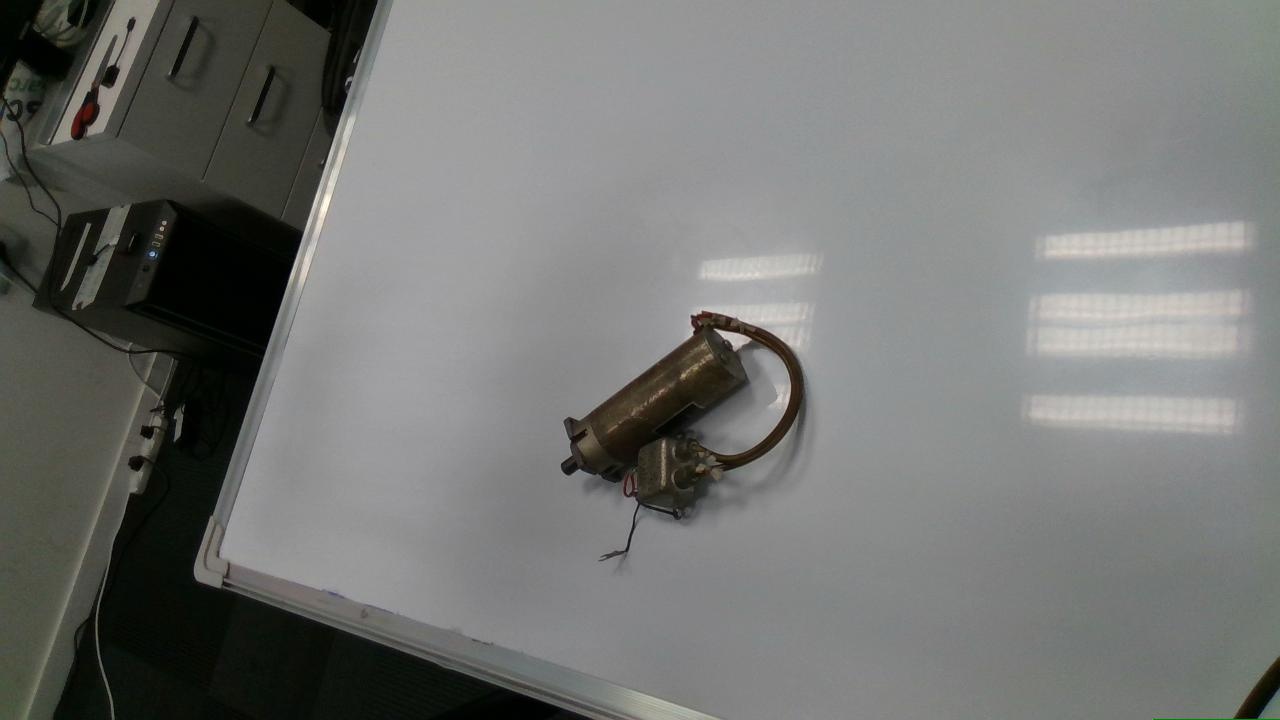}
    \includegraphics[width=.2\linewidth]{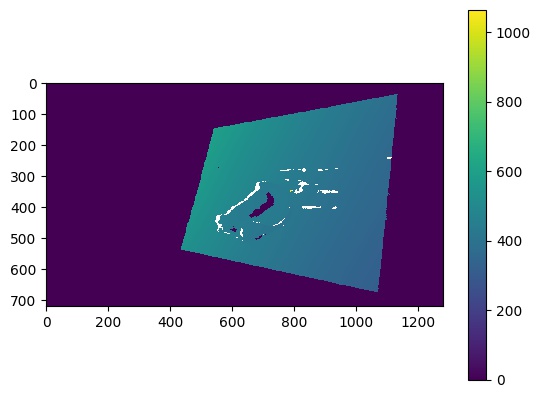}
    \includegraphics[width=.2\linewidth]{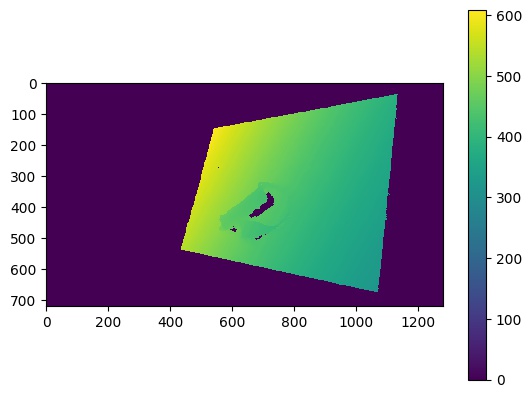}
    \includegraphics[width=.2\linewidth]{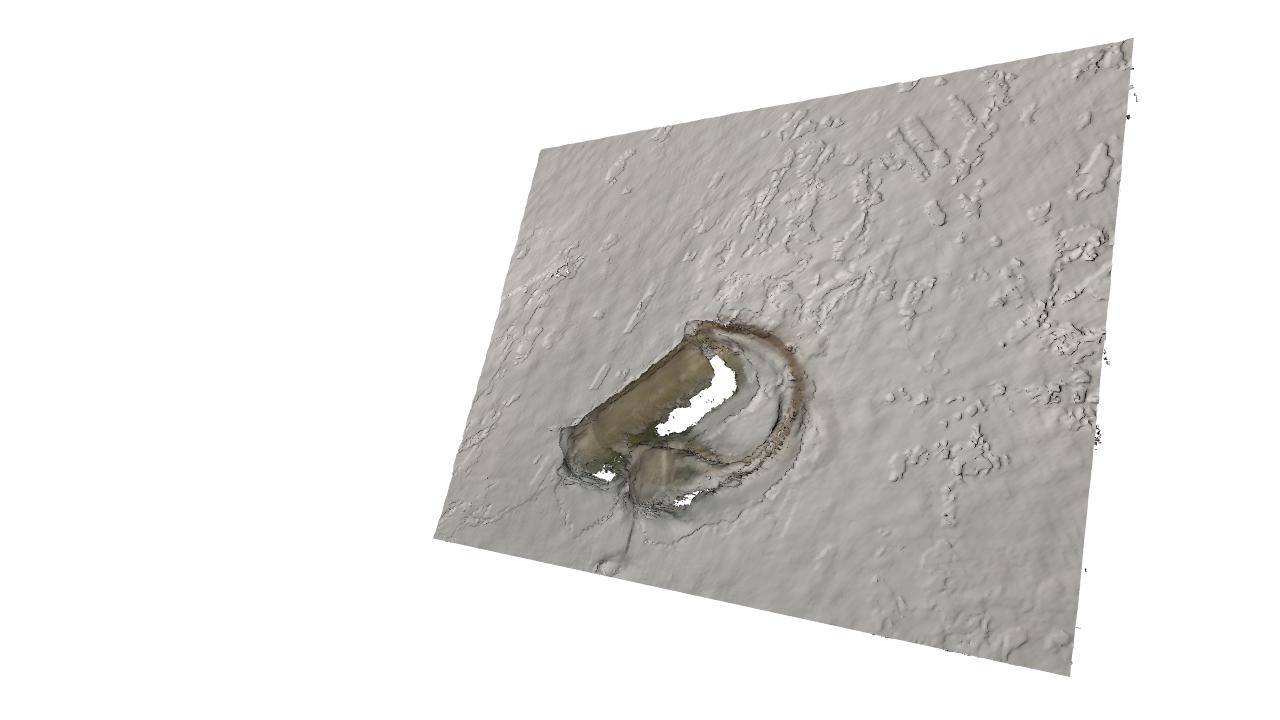}
    
    \includegraphics[width=.2\linewidth]{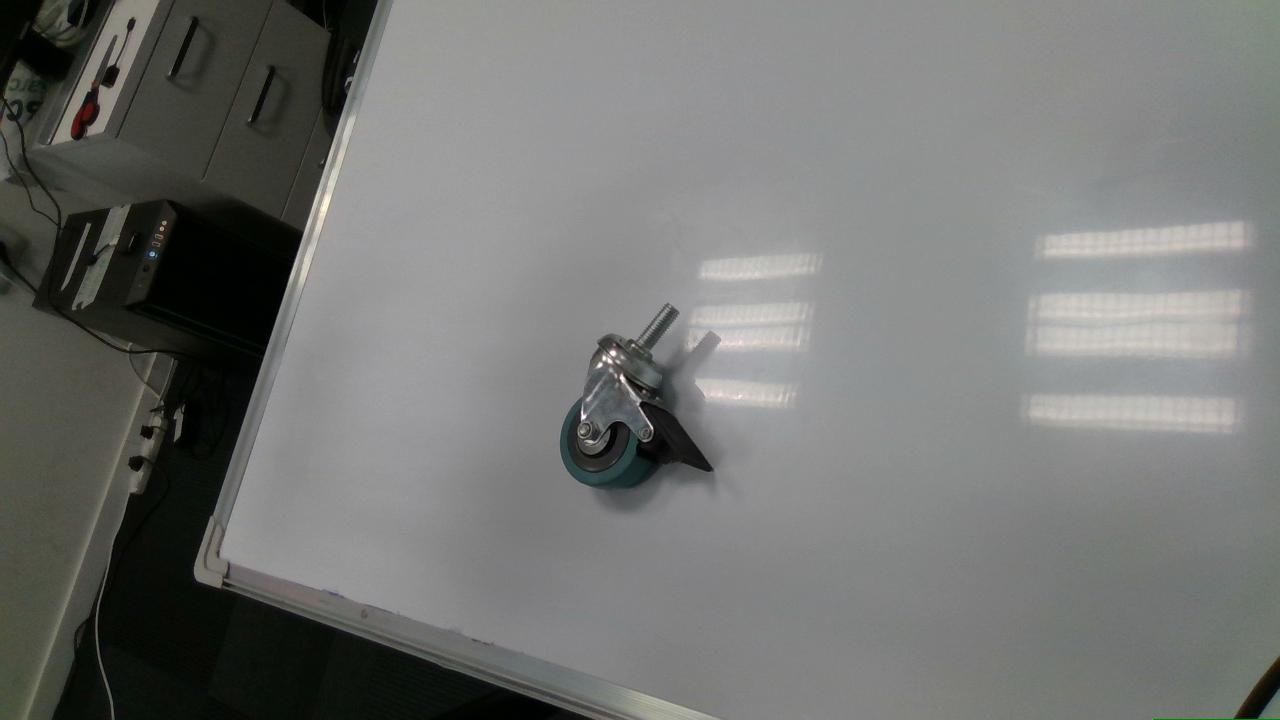}
    \includegraphics[width=.2\linewidth]{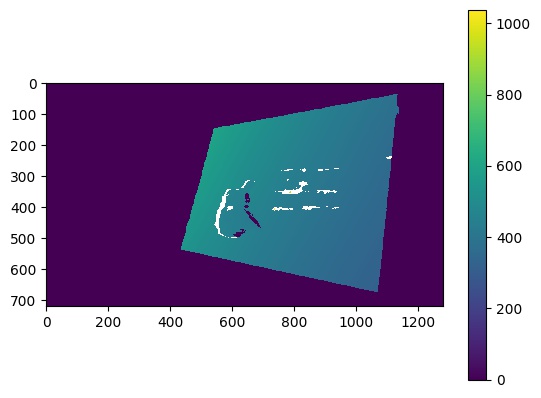}
    \includegraphics[width=.2\linewidth]{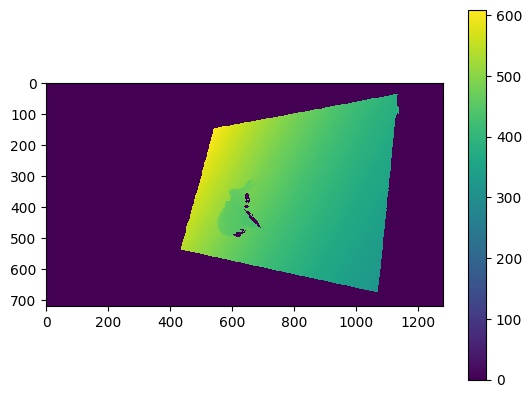}
    \includegraphics[width=.2\linewidth]{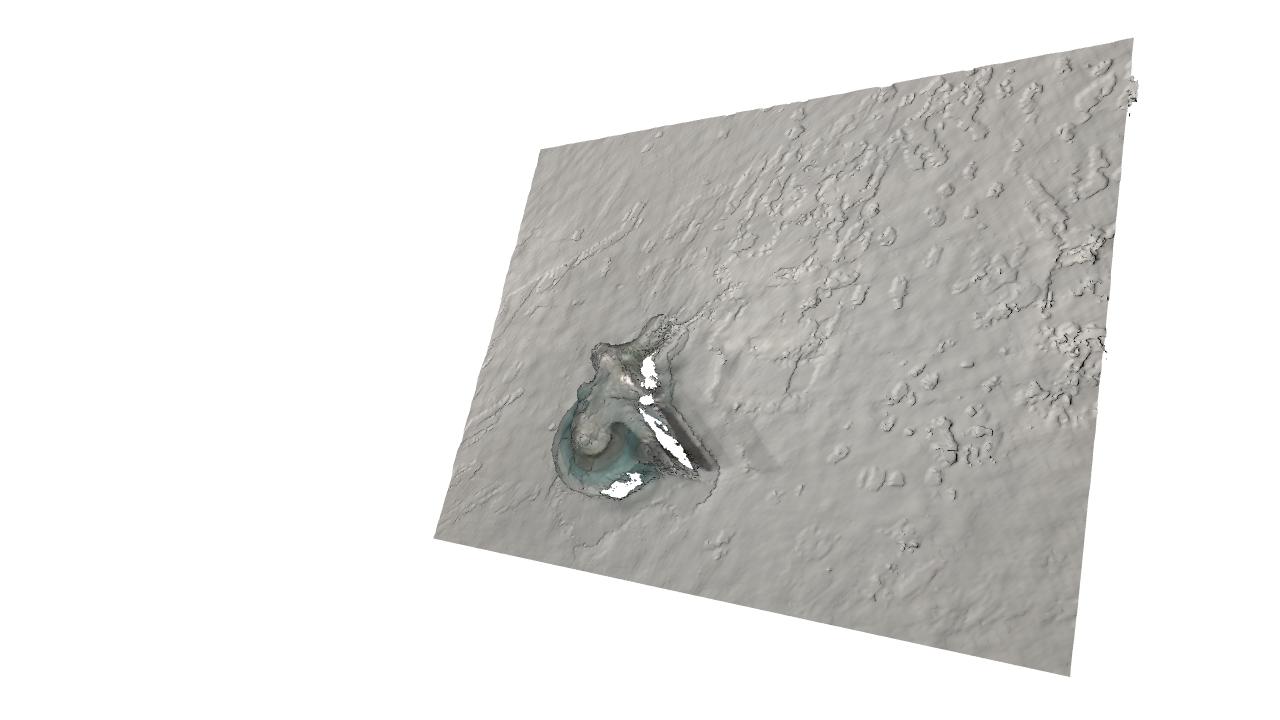}
    
    % Background blue
    \includegraphics[width=.2\linewidth]{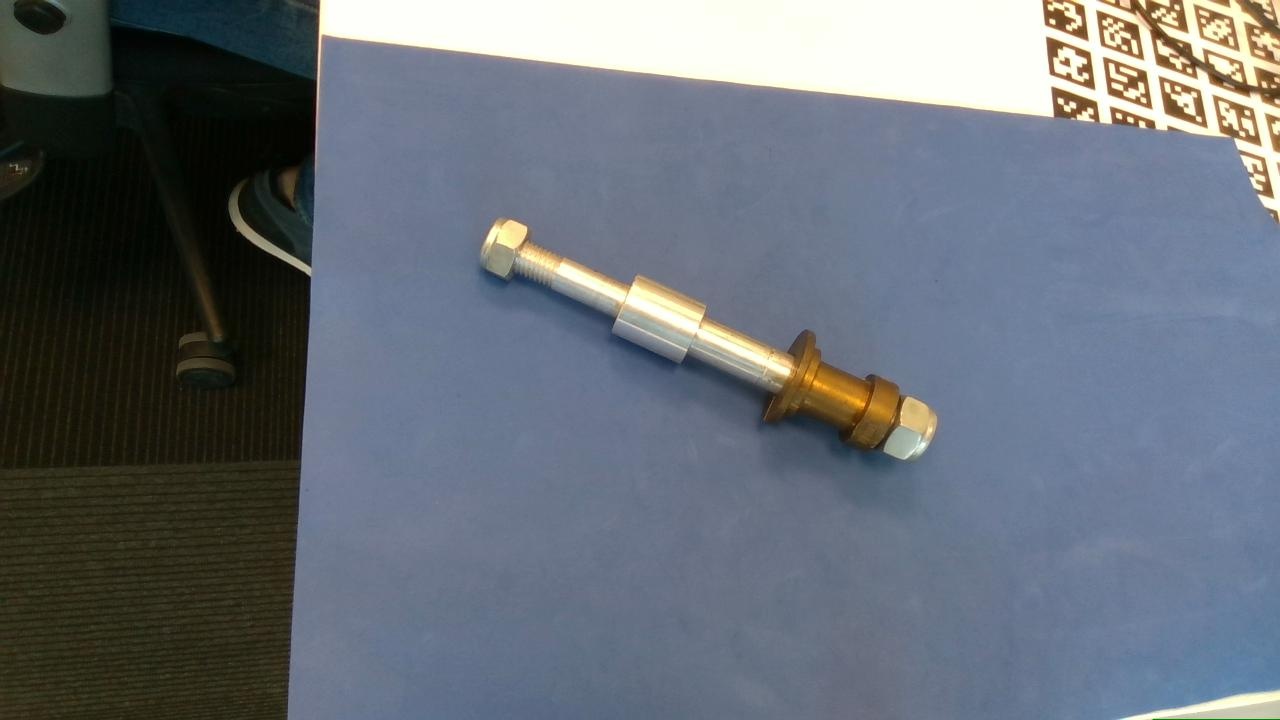}
    \includegraphics[width=.2\linewidth]{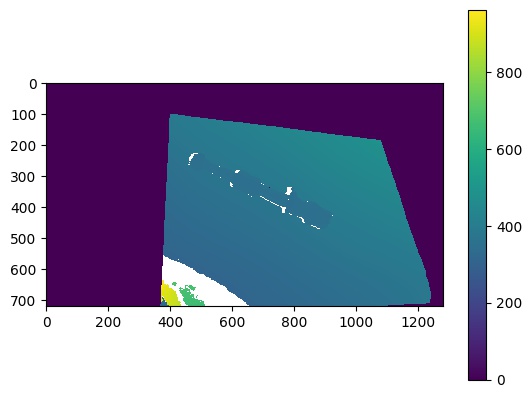}
    \includegraphics[width=.2\linewidth]{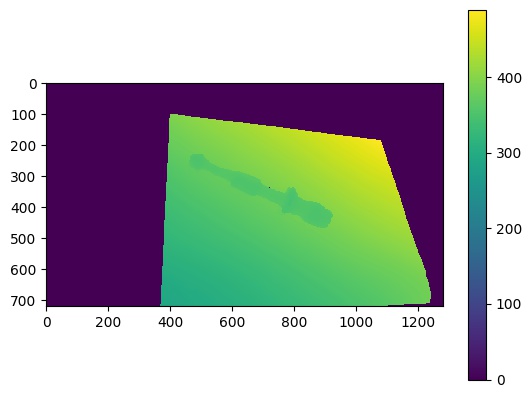}
    \includegraphics[width=.2\linewidth]{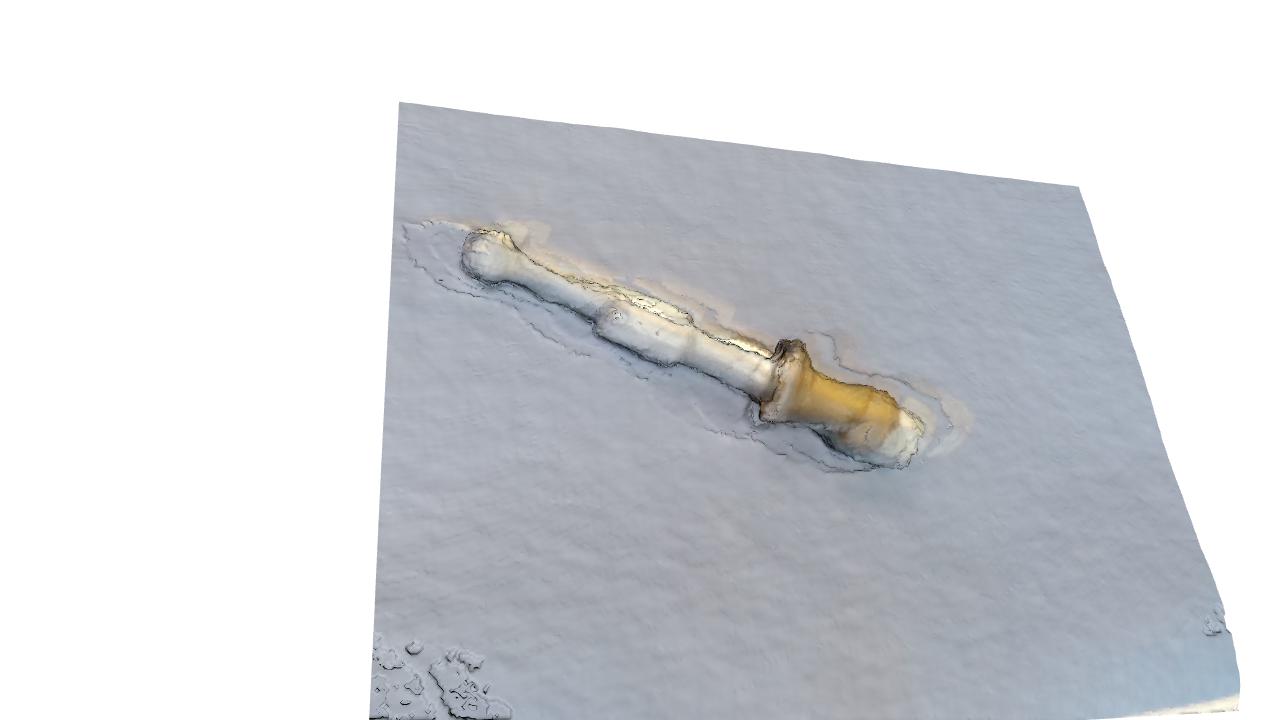}
    
    \includegraphics[width=.2\linewidth]{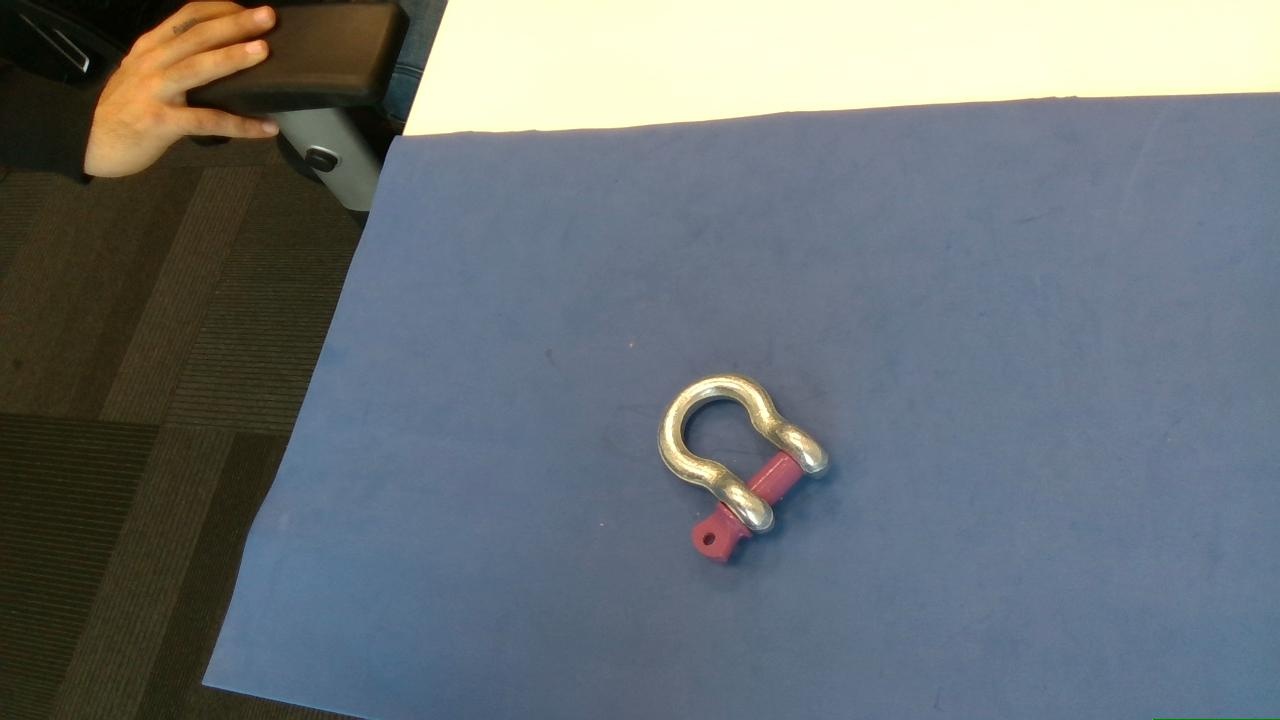}
    \includegraphics[width=.2\linewidth]{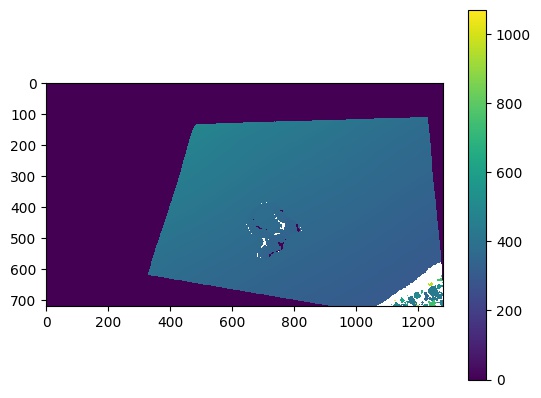}
    \includegraphics[width=.2\linewidth]{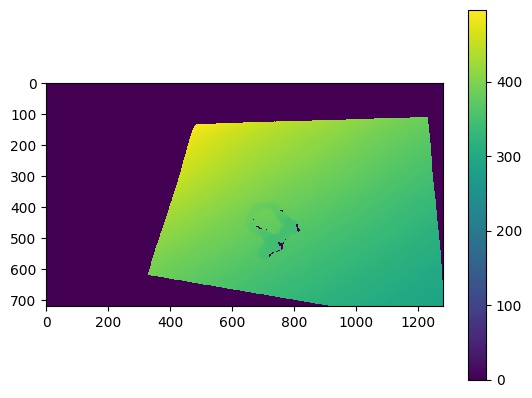}
    \includegraphics[width=.2\linewidth]{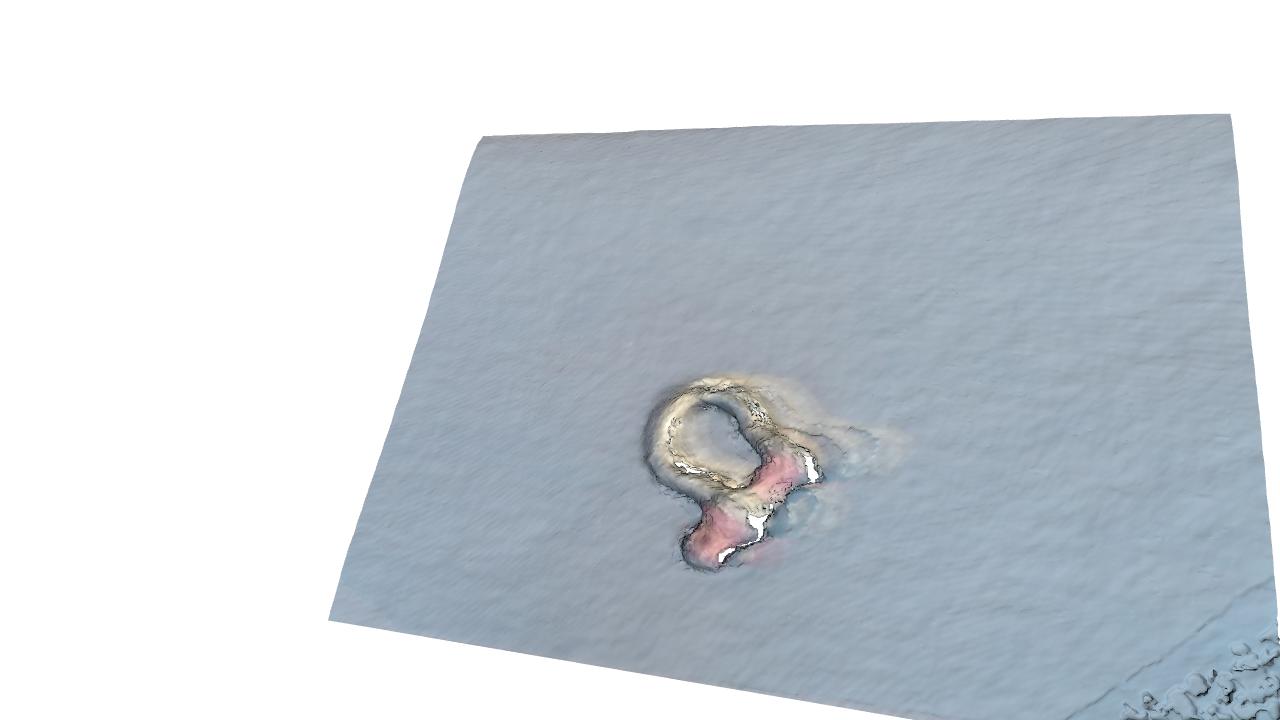}
    
    \includegraphics[width=.2\linewidth]{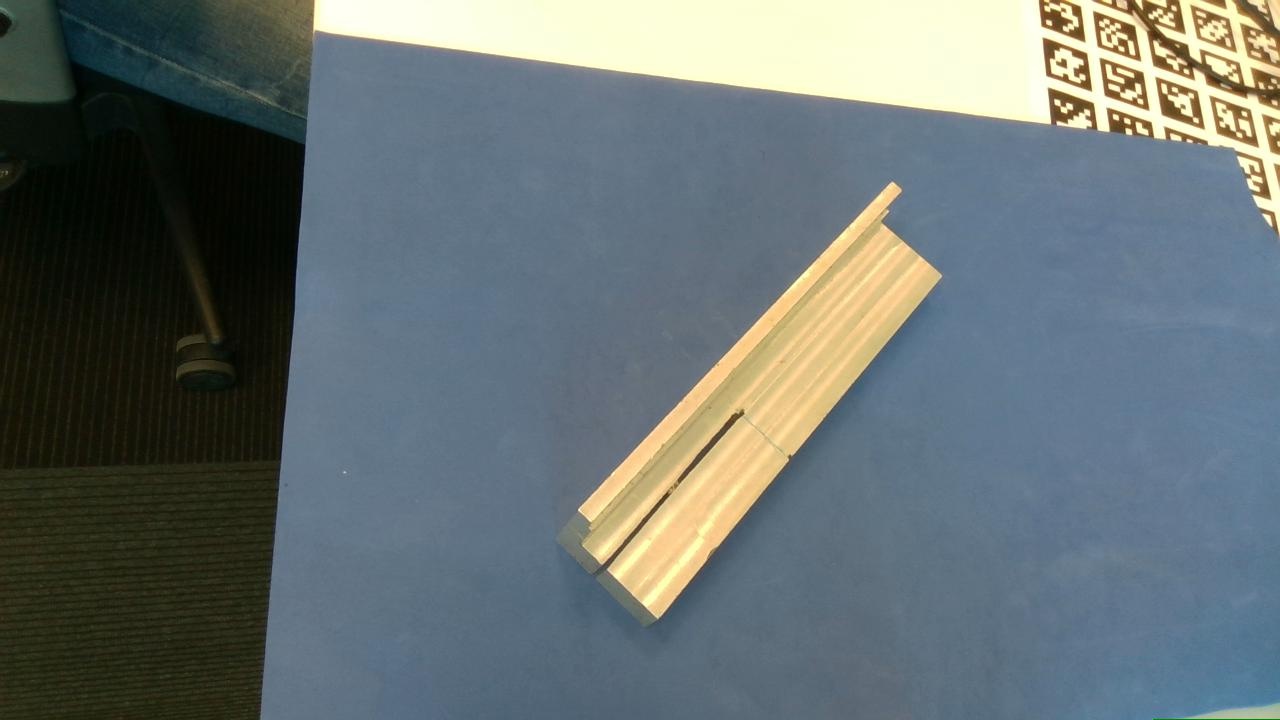}
    \includegraphics[width=.2\linewidth]{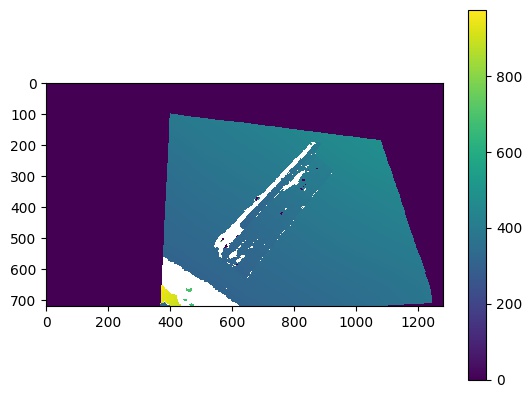}
    \includegraphics[width=.2\linewidth]{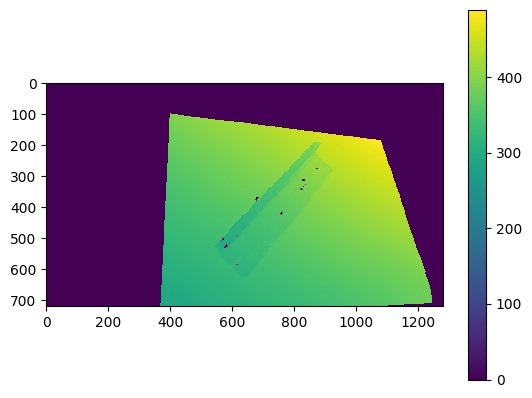}
    \includegraphics[width=.2\linewidth]{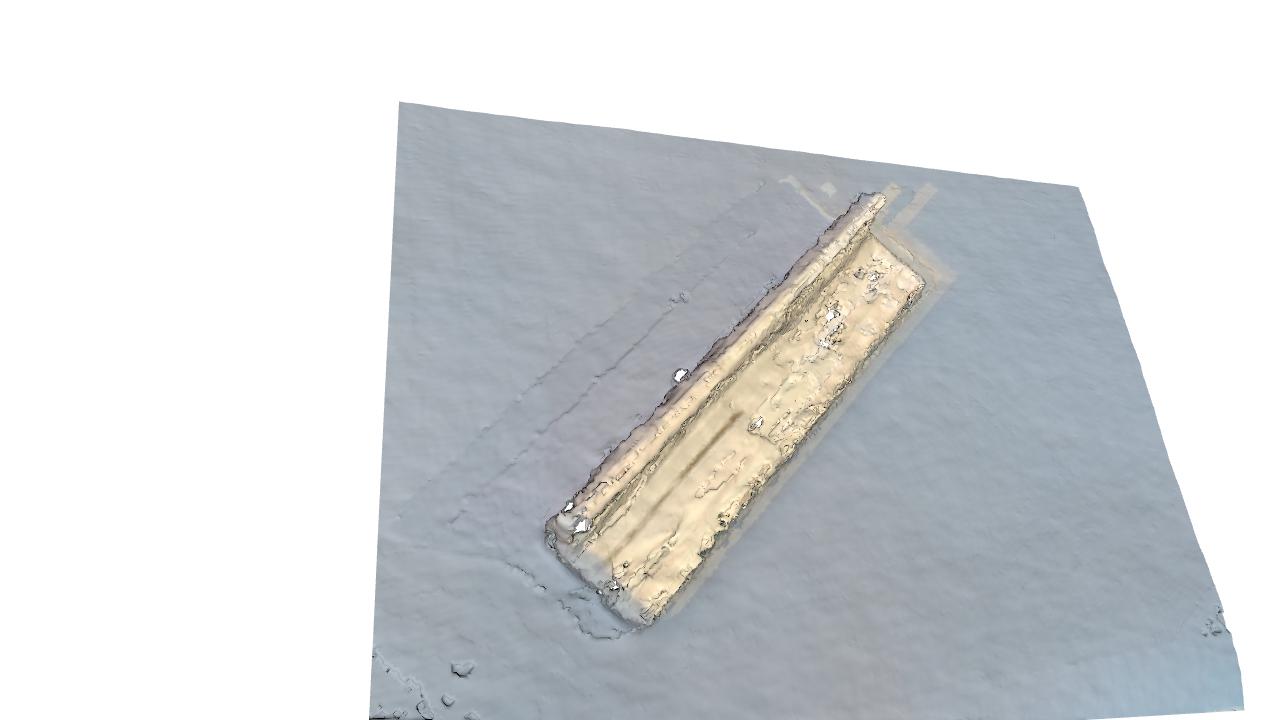}
    
    % Background metal
    \includegraphics[width=.2\linewidth]{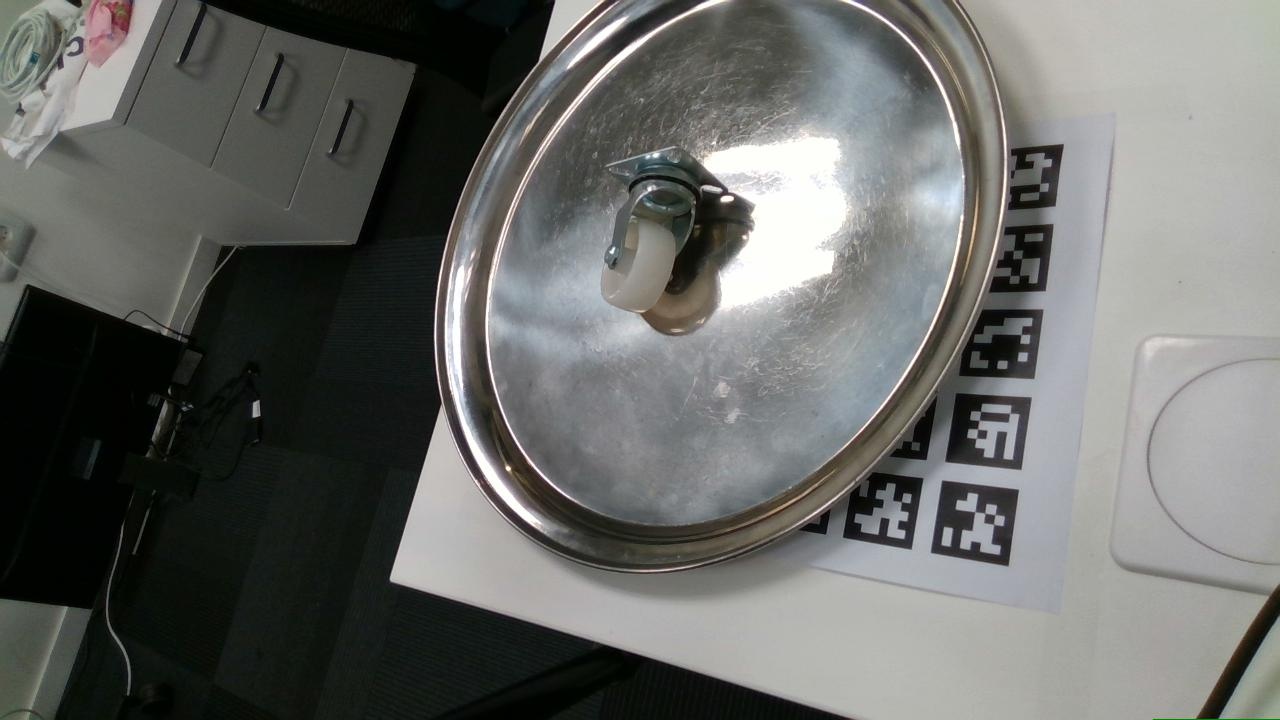}
    \includegraphics[width=.2\linewidth]{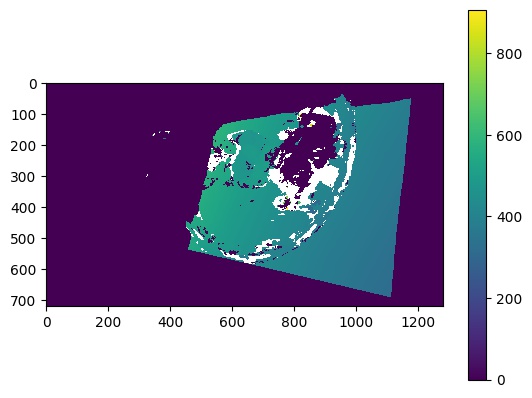}
    \includegraphics[width=.2\linewidth]{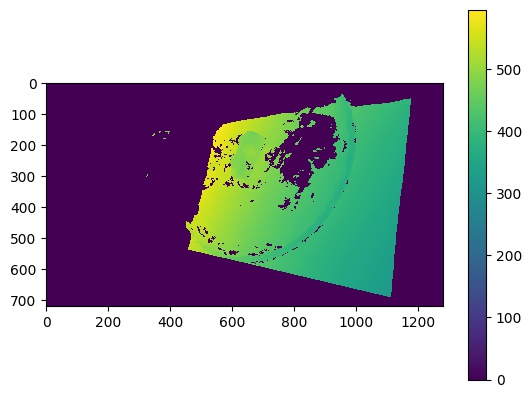}
    \includegraphics[width=.2\linewidth]{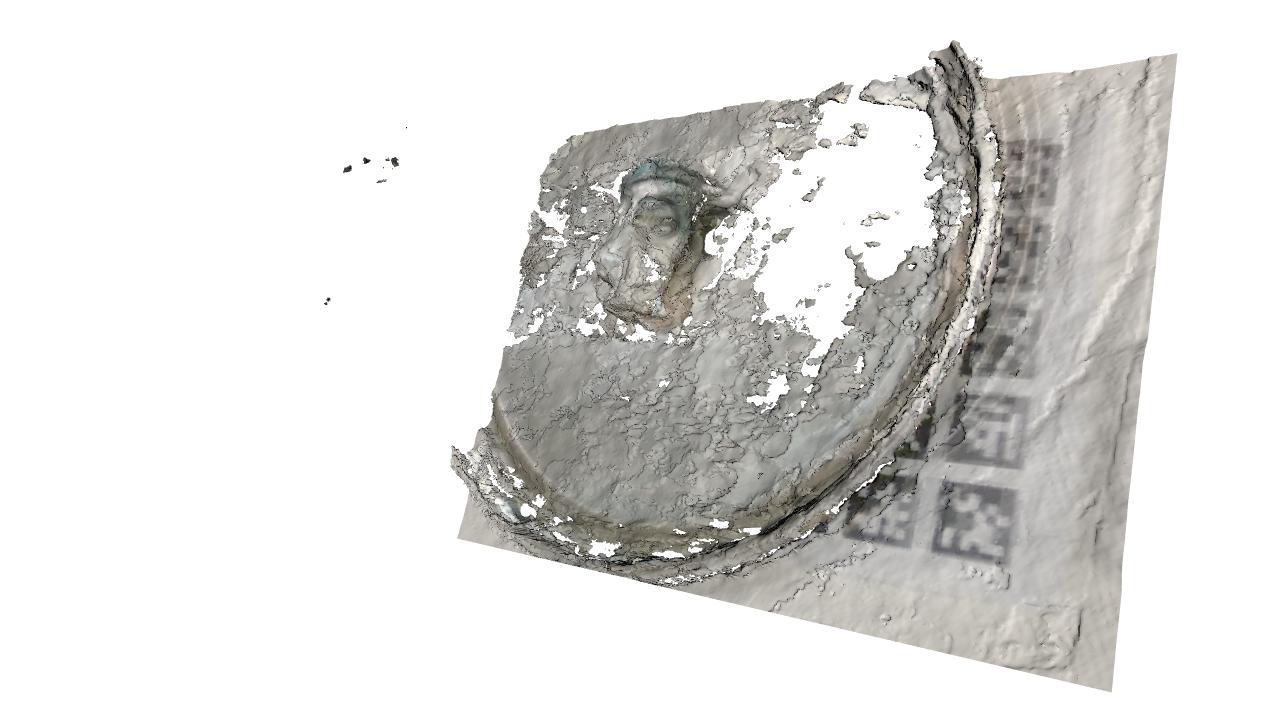}
    
    \includegraphics[width=.2\linewidth]{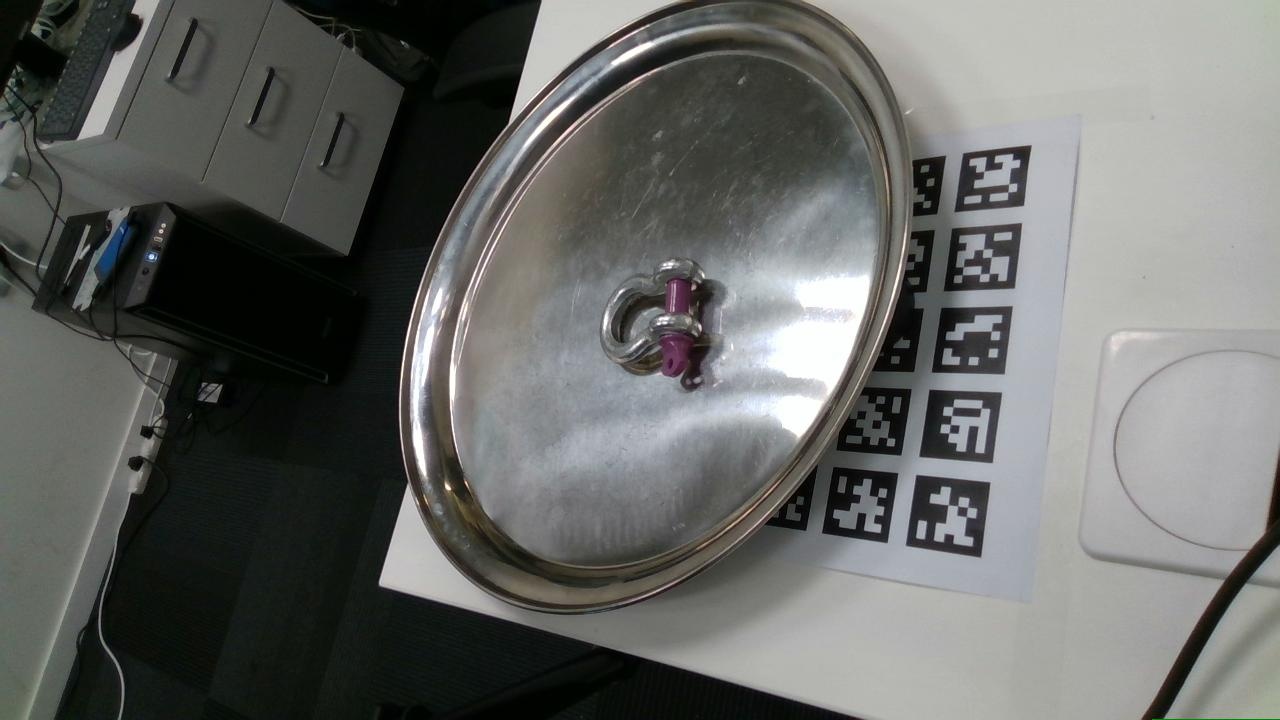}
    \includegraphics[width=.2\linewidth]{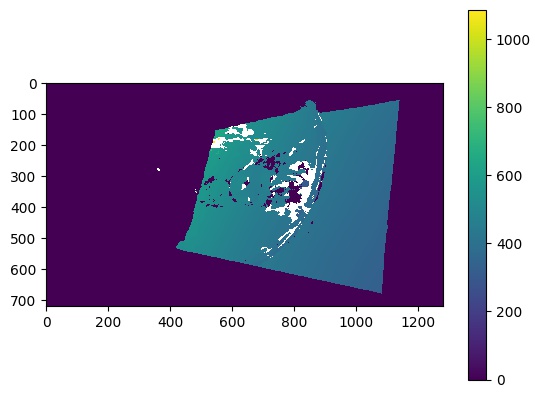}
    \includegraphics[width=.2\linewidth]{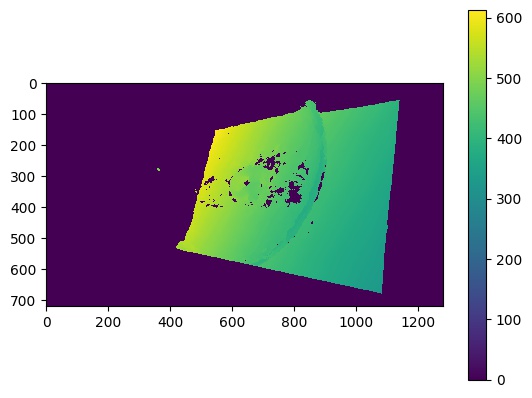}
    \includegraphics[width=.2\linewidth]{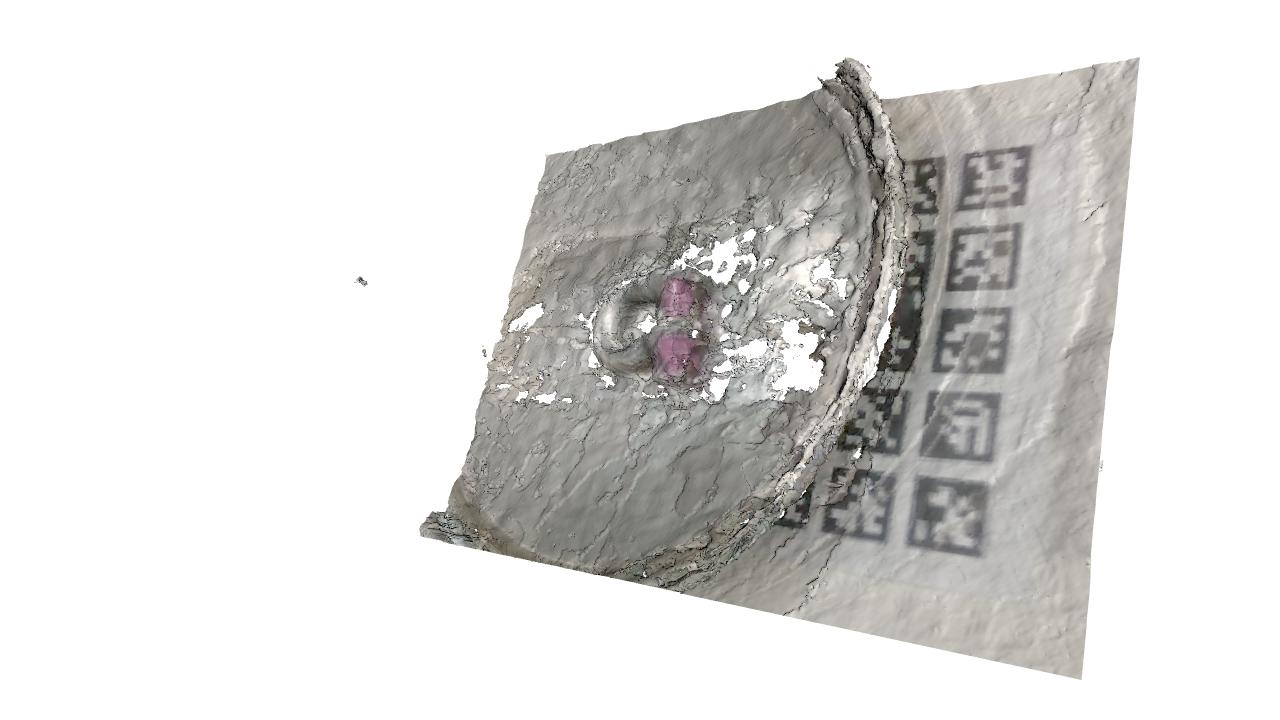}
    
    \includegraphics[width=.2\linewidth]{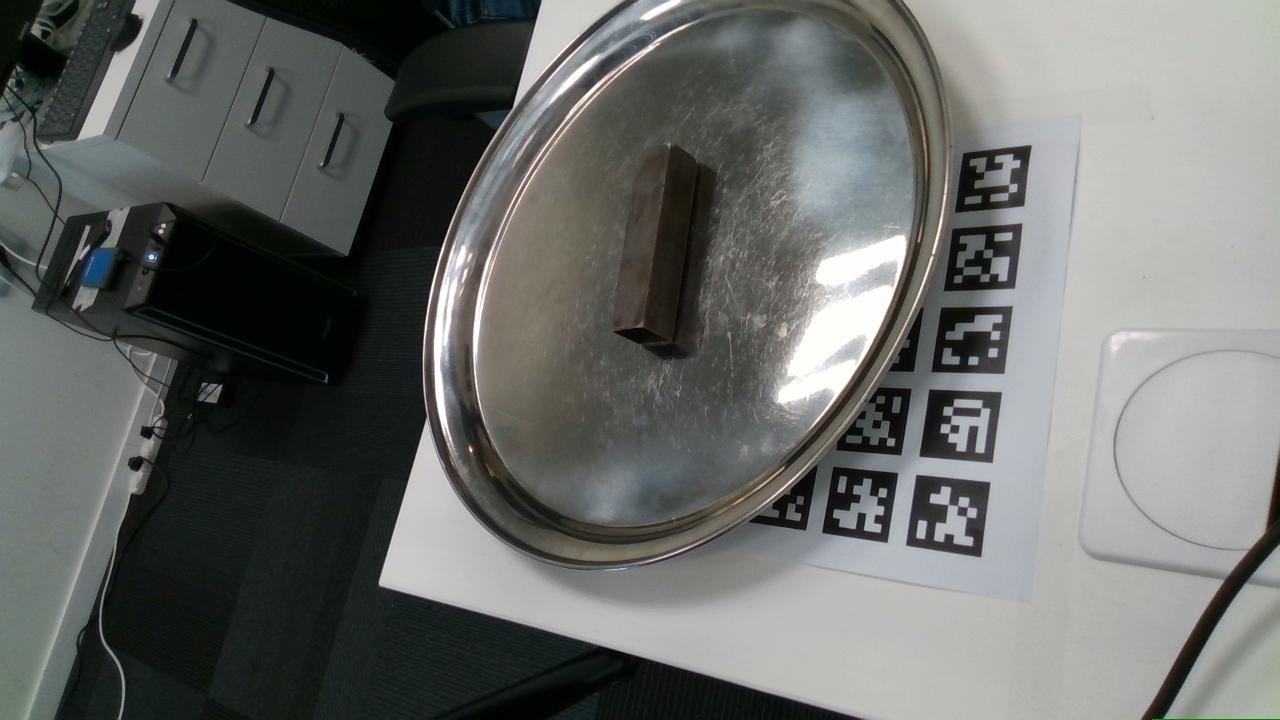}
    \includegraphics[width=.2\linewidth]{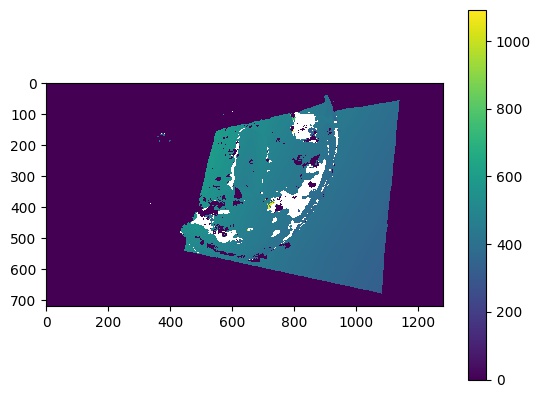}
    \includegraphics[width=.2\linewidth]{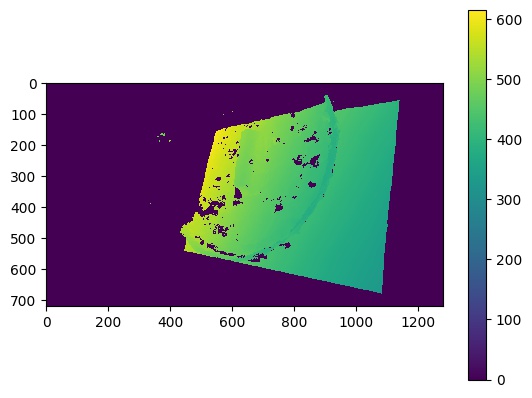}
    \includegraphics[width=.2\linewidth]{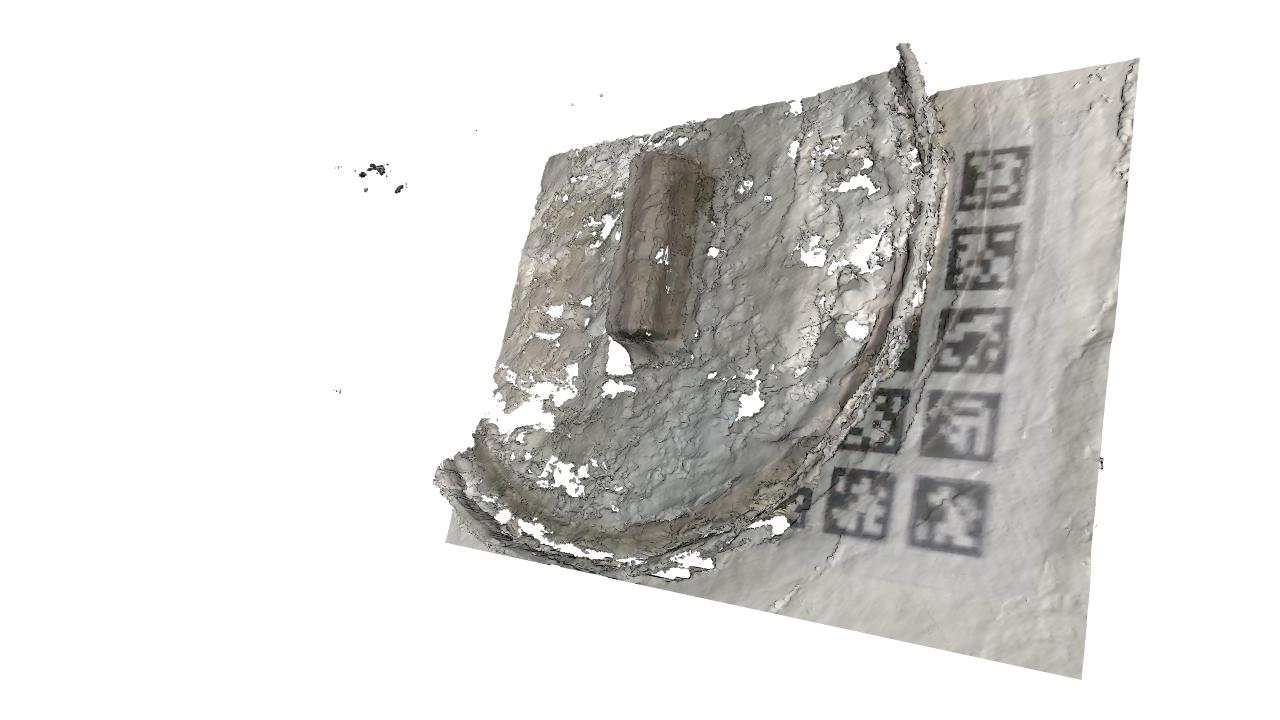}
    
    \caption{\small Samples from our dataset. From left to right: Input RGB, input (measured) depth, fused depth (ground truth), reconstructed mesh (not directly used in training). White "holes" in the measured depth are filled thanks to the depth fusion from multiple views. Zero values in the depth maps correspond to missing pixels which the fusion did not allow to fill. Clearly, the metal background examples (bottom rows) are the more challenging ones, still - some of the missing values are filled. }
    \label{fig:samples}
\end{figure}

\subsection{Data Collection}\label{subsec:data-collection}

We collected depth measurements of 45 industrial and 7 domestic objects, against 3 backgrounds, with 4 or 5 camera viewpoints and 4 poses for each item; yielding about 3,500 data samples. A few samples from the dataset are displayed in \Figref{fig:samples}.

The camera rig was calibrated at the beginning of each data collection session, using an AprilTags matrix \cite{krogius2019iros}. Our registration algorithm applies tag detection, extracts the depth at the tag centers, and then uses the a priori knowledge of the tag matrix to fit the calibration target with a 3D coordinate system and ensure the quality of the depth data.

% \yf{TODO: rephrase}
The measured depth data was particularly poor for the metallic background, due to ambient lighting reflections. In this case, we employed a trick to improve our data, and collected the data at two lighting levels without moving the objects; RGB is always taken from the bright conditions while depth is taken according to usage context either from the dark conditions (in which case it suffers less from ambient reflections or from the bright conditions). 

The ground-truth depth is generated by fusing the information from four camera view-points, using an implementation of TSDF \cite{Curless96accgit} by Zeng et al.~\cite{zeng20163dmatch}, followed by re-projection to each of the original camera view points. Under this data-collection scheme, our depth inputs preserve real-world issues, while the ground-truth is not perfect but significantly better. We show a few data samples in \Figref{fig:samples}. This approach yields lower quality results than the data collection solutions used in \cite{sajjan2019cleargrasp} but is much easier to apply. Despite our efforts, our ground-truth still has holes, so for practical reasons, the training loss is computed only where the ground-truth is valid.

\subsection{Model Architecture}\label{subsec:architecture}

\begin{figure}
    \centering
    \includegraphics[height=5cm]{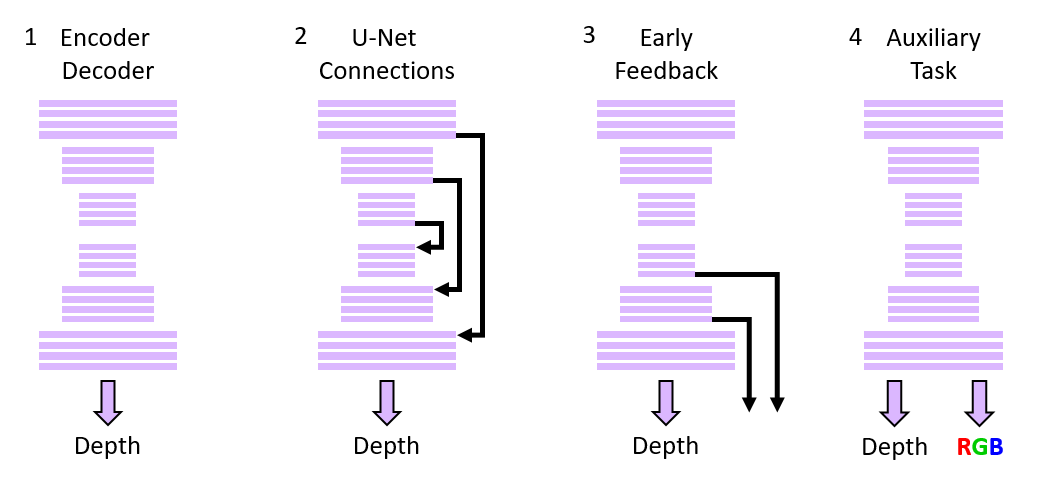}
    \caption{Architectures examined in our work}
    \label{fig:architecture}
\end{figure}

Our deep model is an assemblage of ``best practices" for deep computer vision. We used an encoder-decoder architecture \cite{Hinton504} (\Figref{fig:architecture}.1; our implementation is based on  \cite{Ma18icra}), which is a common choice for dense outputs \cite{he2017mask} and a prevailing architecture for depth estimation (see \Secref{sec:related_work}). 

\paragraph{Backbones} \hfill \\ 
We experimented with ResNet \cite{he2016deep} and VGG \cite{simonyan2014very} backbones; As the latter gave lower accuracy, we exclude it from further discussion. 

\paragraph{Improving the Sharpness of Predicted Depth} \hfill \\
Output blurriness is a known issue of encoder-decoder architectures (it was also shown to be inherent to sparse depth completion by \cite{Ma16iros}, however for a different setting of much sparser inputs). To mitigate that, we applied two modifications to the baseline model: U-Net \cite{ronneberger2015u} -style connections between corresponding encoder and decoder layers of the same resolution (see \Figref{fig:architecture}.2; also commonly used in depth prediction, e.g.~\cite{Garg16eccv,Bloesch18cvpr,Eigen14nips,Ma19icra,Li17iccv,Zhou17cvpr}) and replacing de-convolution layers with max-unpooling for upsampling in the decoder \cite{li2018closedform_FastPhotoStyle}. 

\paragraph{Early Feedback and Auxiliary Outputs} \hfill \\
To expedite training, we used early feedback \cite{GoogLeNet}, i.e. added loss terms for predictions of down-sampled depth at internal decoder layers (\Figref{fig:architecture}.3). To enhance the accuracy of the model, we explored Multi-Task-Learning (MTL) \cite{ruder2017overview,Kendall18cvpr,Eigen15iccv,Cadena16rss} and added an auxiliary task, RGB prediction, or reconstruction (\Figref{fig:architecture}.4). It has been reported in MTL literature that learning correlated tasks can give rise to synergy and higher performance than learning each single task separately. 
% \yf{[Prediction of RGB and Depth are highly correlated in the sense that they tend to share object boundaries, and RGB can provide depth cues such as shading.] this is a serious understatement, considering the vast literature on depth from rgb... consider deleting, or expanding?}. 

\paragraph{Residual Prediction} \hfill \\
The RealSense depth camera and its alternatives come with correction algorithms to improve their raw depth acquisition. This post-processing step usually gives gross but reasonable estimations in our (industrial) scenarios of interest. Acknowledging these capabilities, we also experimented with restating the problem as residual prediction. Formally, denoting input RGB frame as $I$ and post-processed depth as $\tilde{d}$, we fit our model $f$ to predict a correction ("delta") to the input depth, so that final depth prediction $\hat{d}$ is expressed as:
\begin{gather}
    \hat{d} = \tilde{d} + f(I, \tilde{d}).
\end{gather}
In some sense residual-prediction is an easier task conceptually and numerically, since the output lies in a smaller dynamic range. Residual learning is emerging as a promising approach in control and reinforcement learning \cite{zeng2019tossingbot}, where it is used to learn corrections for a parametric control policy. 

\paragraph{Validity Mask} \hfill \\
Residual prediction is supported by appending an additional input channel, quantifying the pixel-wise confidence for each depth value. In our case, we found it practical to use binary values, valid-invalid, but the confidence values could be continuous in the general case.

\subsubsection*{}
The Encoder-Decoder with U-Net connections (\Figref{fig:architecture}.2) was the basis for a model search and experimentation with different combinations of the other features mentioned in the above subsections. We describe these experiments in \Secref{sec:results:ablation}. 

\subsection{Training}\label{subsec:training}

We trained all models on a mixture of our industrial data and on the NYUv2 indoor data \cite{Silberman:ECCV12} at a 1:1 ratio, while the validation set was slanted towards industrial data (85\% to 15\%). These data domains are rather different, so we (a) scaled the NYUv2 depth values by a constant factor to match our working range ($\sim$1m), and (b) applied a ``sparsifier" \cite{Ma18icra} that simulates measured depth by selectively dropping pixel values. The sparsifier is based on \cite{Ma18icra} which creates holes at low texture regions, a typical issue for passive stereo vision systems. The cameras we used emit structured IR light to overcome this issue, but as they do suffer from holes at occlusion boundaries, we modified the sparsifier to simulate this behavior on high gradient regions. 

\paragraph{Objective Function} \hfill \\

The main loss in our work is prediction accuracy, i.e. the difference between the predicted depth and ground truth. Another useful loss was on the magnitude of the prediction's Laplacian, encouraging smoothness (as developed in \cite{Ma16iros,Ma19icra}). We also experimented with losses on the difference between the gradients of the prediction and the ground-truth but this did not prove out to be beneficial.

The general form of the loss we use is:
\begin{gather}
    L = w_1 L_{depth} + w_2 L_{early} + w_3 L_{rgb}, 
\end{gather}
where $L_{depth}$ comprises penalize difference in depth prediction w.r.t. to ground truth, $L_{early}$ penalizes difference of intermediate depth outputs w.r.t. ground truth depth, and $L_{rgb}$ difference in reconstructed RGB (auxiliary task, \Secref{subsec:architecture}) w.r.t. input frame. $\{w_i\}$ are constant weights chosen via hyperparameter search. Specifically, the depth term is defined as: 
\begin{gather}
    L_{depth}(d, \hat{d}) = w_{p} l_{p}(d, \hat{d}) + w_{g} l_{g}(\nabla d, \nabla \hat{d}) + w_{s} l_{s}(\Delta \hat{d}), 
\end{gather}
where $d$, $\hat{d}$ are the ground truth and predicted depth respectively, and we denote by $\nabla d$ the image of gradients and by $\Delta d$ the Laplacian, i.e. $\Delta d = d_{xx}^2 + d_{yy}^2$, per pixel. The weights $w_p$, $w_g$, $w_s$ corresponding to the prediction, gradient (normals) and smoothness terms, respectively, are again chosen via hyperparameter search. 

The early feedback and auxiliary task terms are defined as: 
\begin{gather}
    L_{early} = l_e(\hat{d}_{early}, d), \text{ and } L_{rgb} = l_{rgb}(\hat{I}, I), 
\end{gather}
for $\hat{d}_{early}$ - intermediate depth outputs, and $\hat{I}$ the predicted RGB image (as before, $d$ is the ground truth depth and $I$ the input RGB frame). 

Different choices are possible and discussed in the literature for the distance metrics ($l_{\{p, g, s, d, rgb\}}$). After experimentation (see next clause), we used $\mathcal{L}_1$ to train models for comparison. 

\paragraph{Hyperparameter Search} \hfill \\

 We conducted a hyperparameter search to determine Learning-Rate (LR), LR decay parameters, weight decay, etc. Our hyper parameter search included different types of distance metrics and their relative weights: $\mathcal{L}_1$, $\mathcal{L}_2$, Huber, Adaptive Huber, and RHuber \cite{Irie19icip}. We found negligible differences between the metrics, so from here on we will only mention results for $\mathcal{L}_1$.

\section{Results}\label{sec:results}
% \ys{need to clarify that the plots are the median line, for 4 runs}
In this section, we compare our model to that of Ma and Karaman~\cite{Ma18icra} (\Secref{sec:results:comparative}) using standard metrics (see \Tblref{tbl:metrics}). Subsequently we present the results of an ablation study (\Secref{sec:results:ablation}) outlined in \Secref{subsec:architecture}. All compared models are trained from scratch on the same training set, which is a mixture of random samples at a 1:1 ratio from the NYU Depth v2 (NYUDv2) \cite{Silberman:ECCV12} and our data. The models are compared on a validation set which is a random mixture preferring our data (85\%) over NYUDv2 (15\%).

\setlength{\tabcolsep}{10pt}
\begin{table}[t]
    \centering
    
    \scalebox{0.85}{
        \begin{tabular}{lcl}
             \hline\noalign{\smallskip}
             Metric & Definition & Full Name   \\
             \noalign{\smallskip}
             \hline
             \noalign{\smallskip}
             $\operatorname{RMSE}$ & $\sqrt{\frac{1}{N}\sum_i ( d_i-\hat{d}_i )^2}$  & Root Mean Square Error \\
                \noalign{\smallskip}
             $\operatorname{MAE}$ & $\frac{1}{N}\sum_i \lvert d_i-\hat{d}_i \rvert $ & Mean Absolute Error \\ % Abs. Diff.  
                \noalign{\smallskip}
             $\operatorname{Rel.}$ & $\frac{1}{N}\sum_i \lvert d_i-\hat{d}_i \rvert / d_i $ & Relative Absolute Error \\ % (Abs. Rel.) \cite{Ma18icra},\cite{Ma19icra})
             \noalign{\smallskip}
             \hline
        \end{tabular}
    }
    \caption{\small Evaluation metrics. The sums exclude invalid ground truth pixels. We use $d$ and $\hat{d}$ to denote the target (ground truth) and predicted depth, respectively. The index $i$ runs over $N$, the total number of valid (non-zero) ground truth pixels. }
    \label{tbl:metrics}
\end{table}

\subsection{Comparative Study}
\label{sec:results:comparative}

\begin{figure}
    \centering
    
    \scalebox{0.85}{
        \begin{tabular}{lccc}
            % Header
            \hline\noalign{\smallskip}
            Model     & RMSE [m]  & MAE [m]   & Rel. \\
            \noalign{\smallskip}
            \hline
            \noalign{\smallskip}
            % Body
            Input     & 0.226 & 0.226 & 0.119 \\
            Ma \& Karaman  & 0.052 & 0.039 & 0.092 \\ % their HP
            Ma \& Karaman* & 0.016 & 0.011 & 0.025 \\ % our HP
            Ours      & 0.012 & 0.004 & 0.009 \\ % best-clean
            \hline
        \end{tabular}
    }
    \captionof{table}{\small Validation errors. 1st row: errors in raw inputs w.r.t. ground truth. 2nd row: Ma and Karaman \cite{Ma18icra} trained with best hyperparameters from the original paper, and (3rd row) trained with results of our hyperparameter search. 4th row: our best model (see \Secref{subsec:architecture}). }
    \label{table:accuracy}
    
    \vspace{10pt}
    
    \includegraphics[width=.95\linewidth]{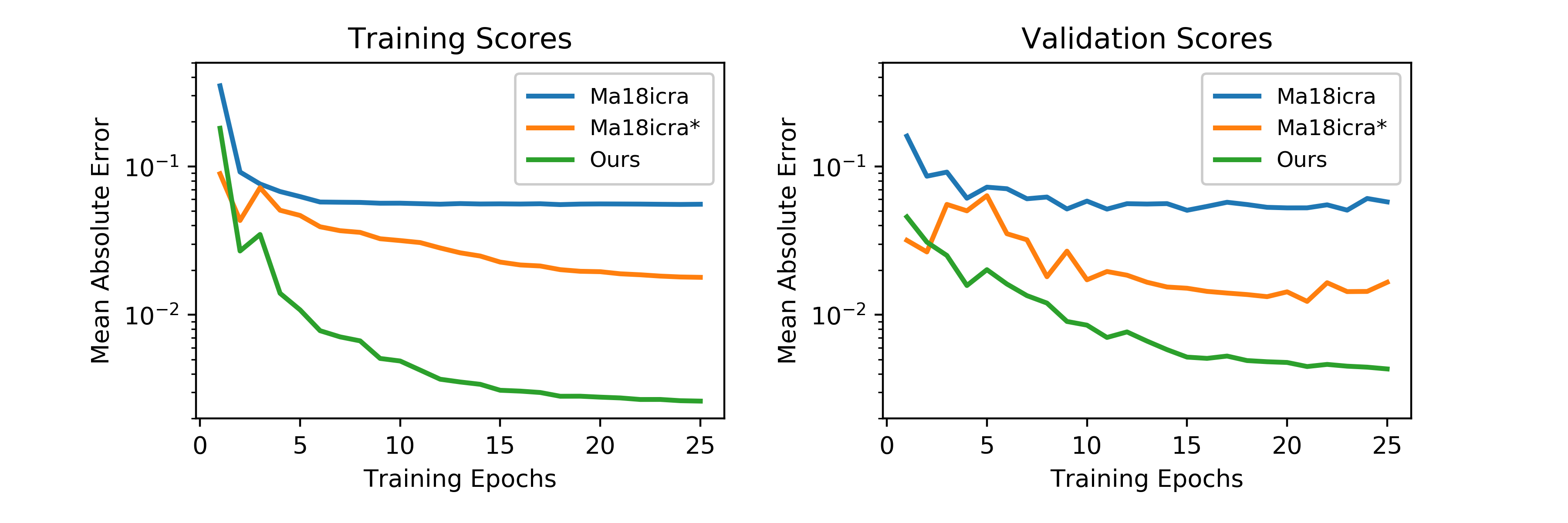}
    \captionof{figure}{Progress of training (left) and validation (right) MAE scores for our best model compared to \cite{Ma18icra}, denoted as Ma \& Karaman 18, after re-training on our data with their original hyper parameters. For fairness, we performed  a hyper parameter search, denoted with an asterisk, which improved their model's performance.}
    \label{fig:model_comparison}
\end{figure}

We comapre our model against Ma and Karaman~\cite{Ma18icra} in \Tblref{table:accuracy}, on the validation set. The 1st row lists the total acquisition error, i.e raw input vs. ground-truth. The 2nd row shows prediction errors of Ma and Karaman's model after training from scratch on our data, using their best hyperparameters. For fair comparison, we performed a hyperparameter search on our data for their model, and show the results on the 3rd row; this was sufficient to reduce their errors by $\sim70\%$. On the 4th row we present our best results, which reduce the errors even further. In \Figref{fig:model_comparison} we depict training errors (MAE, y axis) on the training (left) and validation (right) sets. Notably, we found that our model learns faster and reaches the best accuracy. In the following section, we analyze the individual elements in our model which contribute to this improvement. 

% % \setlength{\tabcolsep}{2pt}

% \end{table}
% \setlength{\tabcolsep}{1.4pt}

% \ys{organize in table above}
%                           rmse  absrel    mae  % mse  
% ground truth              0.226   0.119  0.226  % 0.133  
% vanilla/ma18icra-best     0.052   0.092  0.039  % 0.003  
% vanilla/ma18icra*         0.016   0.025  0.011  % 0.000  
% best/shortcuts2-clean     0.012   0.009  0.004  % 0.000  

% best/shortcuts2           0.001  0.016   0.019  0.009

% incremental/dol           0.001  0.023   0.034  0.015
% incremental/mask          0.001  0.014   0.021  0.009
% incremental/unpool        0.001  0.020   0.032  0.014
% incremental/delta         0.001  0.014   0.016  0.007
% incremental/criterion     0.000  0.017   0.020  0.009
% incremental/baseline      0.000  0.017   0.021  0.010
% decremental/interp        0.005  0.057   0.049  0.022
% decremental/dol           0.000  0.012   0.010  0.004
% decremental/mask          0.000  0.012   0.009  0.004
% decremental/unpool        0.000  0.012   0.010  0.004
% decremental/delta         0.000  0.012   0.014  0.006
% decremental/criterion     0.000  0.011   0.009  0.004
% decremental/delta-interp  0.000  0.013   0.015  0.007

% \yf{are these units correct? MAE/RMSE [m], MSE [m**2], AbsRel [None] (like percents, before multiplication by 100) yes}
% MAE: 0.226
% MSE: 0.133
% RMSE: 0.226
% ABSREL: 0.119

\begin{figure}
    \centering
    \includegraphics[width=\linewidth]{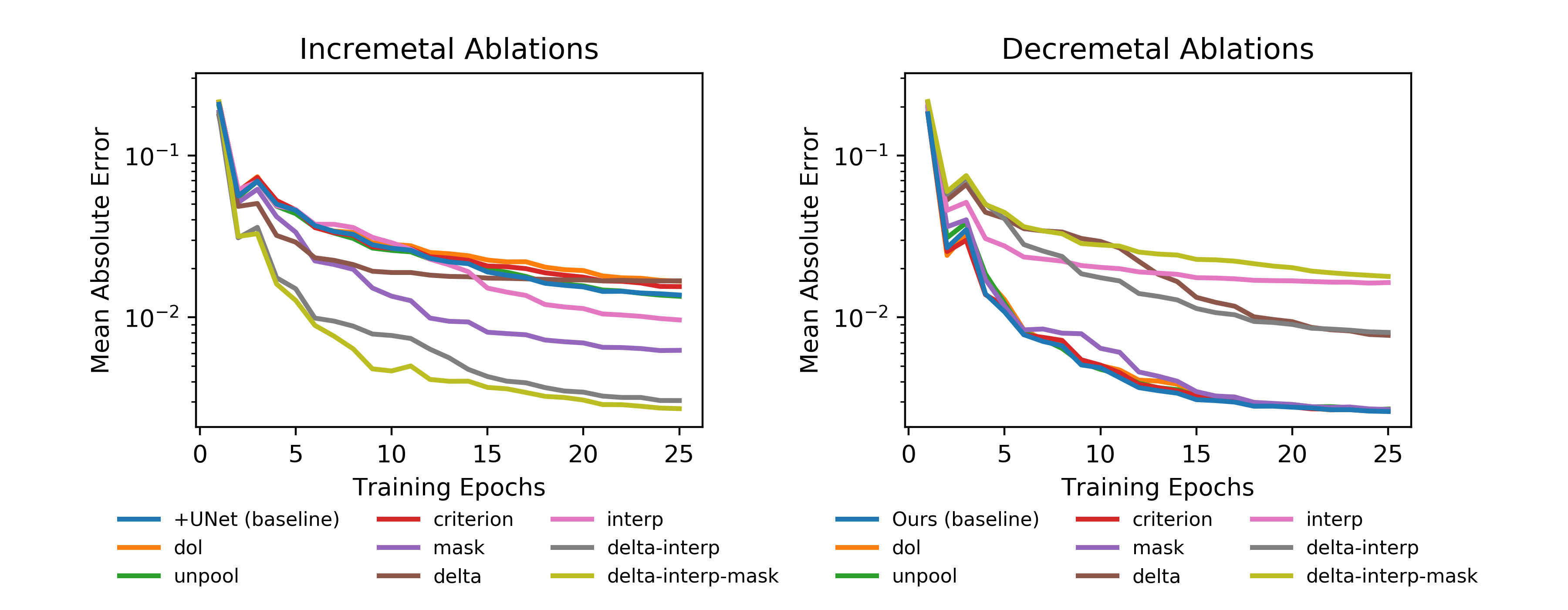}
    \caption{Training scores (MAE) for our ablation study. \textbf{Left:} ``incremental" experiments. +\textit{UNet} is the baseline model, i.e.~\cite{Ma18icra} with the added U-connections. In each trial, a single feature is added, i.e. feature significance corresponds to a \emph{reduction} in error. 
    \textbf{Right:} ``decremental" experiments. the baseline is our best model with all features included. For each trial, a single feature is removed, hence feature significance corresponds to an \emph{increase} in error. See further clarifications in \Tblref{tbl:ablations}, in regard to the modifications from \Secref{subsec:architecture} and \Secref{subsec:training}. For discussion and analysis see \Secref{sec:results:ablation}. }
    \label{fig:ablations}
\end{figure}
% \yf{TODO: split to two subfigures to be able to reference them separately. Use clip and trim properties of \textbackslash includegraphics}

% \yf{TODO: remove figure subtitles, instead use latex captions (can also be done in pure latex). }

% \yf{TODO: ``decremental ablation" is a tautology.. let's call it ``experiments"?}

\subsection{Ablation Study}
\label{sec:results:ablation}

To assess the contribution of each modification listed in \Secref{subsec:architecture}, we report training results from two sets of experiments. In the first set of experiments (``incremetal"; \Figref{fig:ablations}, left panel) we add one modifications to a baseline model consisting of Ma and Karaman's encoder-decoder with added (``U-Net") skip connections. In the second set of experiments (``decremental"; \Figref{fig:ablations}, right panel) we remove one modification at a time from our best-performing model, with the latter (best-performing model) taken as the baseline for comparison. Due to the high co-dependence of a subset of features we introduces a added two trials with more than one modification. Each of the models in the comparison is trained from scratch with the same training hyperparameters. Each experiment corresponds to a feature addressed in \Secref{subsec:architecture} or in \Secref{subsec:training}. Further details on each experiment are given in \Tblref{tbl:metrics}.

\setlength{\tabcolsep}{2pt}
\begin{table}[t]
    \centering
    
    \scalebox{0.85}{
        \begin{tabular}{lll}
             \hline\noalign{\smallskip}
             Experiment Name & Incremental & Decremental   \\
             \noalign{\smallskip}
             \hline
             \noalign{\smallskip}
             
             $+$Unet & \cite{Ma18icra} with added U-connections & {\centering \textendash} \\  \noalign{\smallskip}
             criterion & Criterion from \Secref{subsec:training}  & $\mathcal{L}_1$ between predicted depth and GT  \\  \noalign{\smallskip}
             dol & Use early (depth) feedback  & No early feedback \\ \noalign{\smallskip}
             mask & Use binary validity mask  & No validity mask input \\ \noalign{\smallskip}
             unpool & Use max-unpooling  & Use deconvolution layers \\ \noalign{\smallskip}
             delta & Residual prediction, input interpolation & Depth prediction \\ \noalign{\smallskip}
             interp & \textendash & Predict depth, with input interpolation \\ \noalign{\smallskip}
             delta-interp & \textendash  & Predict depth, no input interpolation \\ \noalign{\smallskip}
             delta-interp-mask & Residual, interpolation, validity mask & \textendash \\
             \noalign{\smallskip}
             \hline
        \end{tabular}
    }
    \caption{\small Ablation experiments, correspondence to \Secref{subsec:architecture}. Experiment name is as appears in \Figref{fig:ablations}. Incremental column lists corresponding modification(s) to baseline for incremental experiments (\Figref{fig:ablations} left). Decremental column lists corresponding modifications to baseline for decremental experiments (\Figref{fig:ablations} right).  }
    \label{tbl:ablations}
\end{table}

The incremental experiments (\Figref{fig:ablations}, left panel) show that a noted improvement was obtained w.r.t. the baseline model by adding a validity mask input channel. Not surprisingly, the most significant improvement was obtained by interpolating depth measurements input to the model and predicting the residual, while the rest of the modifications do not seem to have a significant effect on their own. Residual prediction over interpolated input and validity mask accounts for most of the improvement w.r.t. the baseline. The same behavior can be seen in the decremental experiments (\Figref{fig:ablations}, right panel), where performance was affected by removing residual prediction, i.e. training to directly predict depth instead. Interestingly, removing the residual prediction reduced accuracy (``delta", ``delta-interp" experiments) even when model was still input with interpolated depth (``delta"). However, predicting residual depth from non-interpolated inputs caused an even more severe degradation (``interp" experiment).

\section{Conclusions and Future Work}\label{sec:conclusions}

% recap on data
In this paper, we provide a novel industrial objects dataset of RGBD images and a method to complete missing depth points due to reflections of specular objects. The main limitations of the dataset are:
\begin{itemize}
    \item Variability: 45 objects, 3 backgrounds.
    \item Size: {$\sim$}3,500 samples, i.e. different combinations of object pose ({$\sim$}4 per object) and four static camera view-points.
    \item Fidelity: the ground truth depth suffers from holes.
\end{itemize}

% automatic robotic data collection
A compelling avenue for future work is automating the data collection using a robot mounted camera. This will make it easier to acquire a larger variety of objects, increase the number of view-points, and open interesting secondary research questions. With a camera-in-hand, one can generate the ground-truth, either by applying multiple-camera geometry on the robot's poses or by Structure-from-Motion techniques. The efficiency of this processes could benefit from applying \emph{Active Perception} principles \cite{bajcsy1992active,chen2011active,bajcsy2018revisiting}. Moreover, we expect active perception to improve data fidelity, as \cite{wu2015active} has demonstrated in the domestic domain, since different view angles modify the perceived reflections and may completely eliminate them. This will also allow us to do surface normals prediction, which currently is problematic with the data we have at hand. Additionally, the robot will manipulate the objects and randomize the scenes to expand variability and boost generalization capability of trained models.

% Synthetic data
One alternative to this automation is to generate photo-realistic synthetic data. This has been done for transparent consumer products \cite{sajjan2019cleargrasp} and could be attempted for industrial applications, largely thanks to the availability of large industrial CAD datasets \cite{ABCDataset}. The flexibility of synthetic frameworks is appealing but it must be noted that they require large engineering and artistic-design efforts to render a large variety of scenes and bridge the synthetic-real domain gap. In our case. we have two potential domain-gaps which are slightly different from each other:
\begin{itemize}
    \item Photo-realism: the RGB images need to appear real.
    \item Depth issues: it is straightforward to produce complete synthetic depth for ground-truth, but it is more challenging to imitate the device-specific issues and artifacts that arise in real-world acquisition systems.
\end{itemize}
A practical approach might be to mix synthetic and real data, as well as applying domain-adaptation techniques \cite{da_book}.

% Model improvements

%YS: This felt a bit vague, and repetitive, so I removed it
% "In addition, it was demonstrated that synergies of best practices from similar deep models improve the overall performance."
Notably, the model that we developed performed better than related counterparts, owing to the synergy between the best-practices assembled from previous analyses. The paramount modifications are the additional U-connections and rephrasing the problem in residual terms, that is, predicting a small additive correction on top of a reasonable initial estimation. It might be possible to drive the model performance further by tracking the latest improvements in deep models, e.g. higher capacity ResNet blocks \cite{Res2Net_Gao_2019}. Prior research \cite{Eigen15iccv,sajjan2019cleargrasp} achieved compelling progress by taking surface normals into consideration. Consequently, an interesting continuation of the current research would be to add an auxiliary task of predicting surface-normals from RGB. We note that in this work we refrained from doing so due to lack of data, namely the ground-truth being relatively noisy and scarce. 

Depth completion is a rapidly evolving research topic, but it still lacks clear industrial requirements or benchmarks to test against. 
As a by-pass, we suggest that it would be interesting to gauge the impact of this post-processing step on physical manufacturing tasks, such as object grasping and manipulations. It could be exciting to demonstrate the practical implications of this seemingly tepid depth-completion task. We believe that it is a pivotal enabler for Industry 4.0, with vast consequences to global economy.

\clearpage
% ---- Bibliography ----
%
% BibTeX users should specify bibliography style 'splncs04'.
% References will then be sorted and formatted in the correct style.
%
\bibliographystyle{splncs04}
\bibliography{egbib.bib}

\begin{thebibliography}{10}
\providecommand{\url}[1]{\texttt{#1}}
\providecommand{\urlprefix}{URL }
\providecommand{\doi}[1]{https://doi.org/#1}

\bibitem{bajcsy2018revisiting}
Bajcsy, R., Aloimonos, Y., Tsotsos, J.K.: Revisiting active perception.
  Autonomous Robots  \textbf{42}(2),  177--196 (2018)

\bibitem{bajcsy1992active}
Bajcsy, R., Campos, M.: Active and exploratory perception. CVGIP: Image
  Understanding  \textbf{56}(1),  31--40 (1992)

\bibitem{Bloesch18cvpr}
Bloesch, M., Czarnowski, J., Clark, R., Leutenegger, S., Davison, A.J.:
  Codeslam—learning a compact, optimisable representation for dense visual
  slam. In: Proceedings of the IEEE conference on computer vision and pattern
  recognition. pp. 2560--2568 (2018)

\bibitem{Cadena16rss}
Cadena, C., Dick, A.R., Reid, I.D.: Multi-modal auto-encoders as joint
  estimators for robotics scene understanding. In: Robotics: Science and
  Systems. vol.~7 (2016)

\bibitem{chen2011active}
Chen, S., Li, Y., Kwok, N.M.: Active vision in robotic systems: A survey of
  recent developments. The International Journal of Robotics Research
  \textbf{30}(11),  1343--1377 (2011)

\bibitem{da_book}
Csurka, G.: Domain Adaptation in Computer Vision Applications. Springer
  Publishing Company, Incorporated, 1st edn. (2017)

\bibitem{Curless96accgit}
Curless, B., Levoy, M.: A volumetric method for building complex models from
  range images. In: Proceedings of the 23rd annual conference on Computer
  graphics and interactive techniques. pp. 303--312 (1996)

\bibitem{Eigen15iccv}
Eigen, D., Fergus, R.: Predicting depth, surface normals and semantic labels
  with a common multi-scale convolutional architecture. In: Proceedings of the
  IEEE international conference on computer vision. pp. 2650--2658 (2015)

\bibitem{Eigen14nips}
Eigen, D., Puhrsch, C., Fergus, R.: Depth map prediction from a single image
  using a multi-scale deep network. In: Advances in neural information
  processing systems. pp. 2366--2374 (2014)

\bibitem{faugeras1988motion}
Faugeras, O.D., Lustman, F.: Motion and structure from motion in a piecewise
  planar environment. International Journal of Pattern Recognition and
  Artificial Intelligence  \textbf{2}(03),  485--508 (1988)

\bibitem{Res2Net_Gao_2019}
Gao, S., Cheng, M.M., Zhao, K., Zhang, X.Y., Yang, M.H., Torr, P.H.: Res2net: A
  new multi-scale backbone architecture. IEEE Transactions on Pattern Analysis
  and Machine Intelligence p. 1–1 (2019). \doi{10.1109/tpami.2019.2938758},
  \url{http://dx.doi.org/10.1109/TPAMI.2019.2938758}

\bibitem{Garg16eccv}
Garg, R., BG, V.K., Carneiro, G., Reid, I.: Unsupervised cnn for single view
  depth estimation: Geometry to the rescue. In: European Conference on Computer
  Vision. pp. 740--756. Springer (2016)

\bibitem{Giannone19cvpr_workshop}
Giannone, G., Chidlovskii, B.: Learning common representation from rgb and
  depth images. In: Proceedings of the IEEE Conference on Computer Vision and
  Pattern Recognition Workshops. pp.~0--0 (2019)

\bibitem{godard2017unsupervised}
Godard, C., Mac~Aodha, O., Brostow, G.J.: Unsupervised monocular depth
  estimation with left-right consistency. In: Proceedings of the IEEE
  Conference on Computer Vision and Pattern Recognition. pp. 270--279 (2017)

\bibitem{handa2014benchmark}
Handa, A., Whelan, T., McDonald, J., Davison, A.J.: A benchmark for rgb-d
  visual odometry, 3d reconstruction and slam. In: 2014 IEEE international
  conference on Robotics and automation (ICRA). pp. 1524--1531. IEEE (2014)

\bibitem{he2017mask}
He, K., Gkioxari, G., Dollár, P., Girshick, R.: Mask r-cnn (2017)

\bibitem{he2016deep}
He, K., Zhang, X., Ren, S., Sun, J.: Deep residual learning for image
  recognition. In: Proceedings of the IEEE conference on computer vision and
  pattern recognition. pp. 770--778 (2016)

\bibitem{Hinton504}
Hinton, G.E., Salakhutdinov, R.R.: Reducing the dimensionality of data with
  neural networks. Science  \textbf{313}(5786),  504--507 (2006).
  \doi{10.1126/science.1127647},
  \url{https://science.sciencemag.org/content/313/5786/504}

\bibitem{huang2002motion}
Huang, T.S., Netravali, A.N.: Motion and structure from feature
  correspondences: A review. In: Advances In Image Processing And
  Understanding: A Festschrift for Thomas S Huang, pp. 331--347. World
  Scientific (2002)

\bibitem{Irie19icip}
Irie, G., Kawanishi, T., Kashino, K.: Robust learning for deep monocular depth
  estimation. In: 2019 IEEE International Conference on Image Processing
  (ICIP). pp. 964--968. IEEE (2019)

\bibitem{jrgensen2019monocular}
Jörgensen, E., Zach, C., Kahl, F.: Monocular 3d object detection and box
  fitting trained end-to-end using intersection-over-union loss (2019)

\bibitem{Kendall18cvpr}
Kendall, A., Gal, Y., Cipolla, R.: Multi-task learning using uncertainty to
  weigh losses for scene geometry and semantics. In: Proceedings of the IEEE
  conference on computer vision and pattern recognition. pp. 7482--7491 (2018)

\bibitem{ABCDataset}
Koch, S., Matveev, A., Jiang, Z., Williams, F., Artemov, A., Burnaev, E.,
  Alexa, M., Zorin, D., Panozzo, D.: {ABC:} {A} big {CAD} model dataset for
  geometric deep learning. CoRR  \textbf{abs/1812.06216} (2018),
  \url{http://arxiv.org/abs/1812.06216}

\bibitem{krogius2019iros}
Krogius, M., Haggenmiller, A., Olson, E.: Flexible layouts for fiducial tags.
  In: Proceedings of the {IEEE/RSJ} International Conference on Intelligent
  Robots and Systems {(IROS)} (2019)

\bibitem{kuznietsov2017semi}
Kuznietsov, Y., Stuckler, J., Leibe, B.: Semi-supervised deep learning for
  monocular depth map prediction. In: Proceedings of the IEEE conference on
  computer vision and pattern recognition. pp. 6647--6655 (2017)

\bibitem{Laina16ic3dv}
Laina, I., Rupprecht, C., Belagiannis, V., Tombari, F., Navab, N.: Deeper depth
  prediction with fully convolutional residual networks. In: 2016 Fourth
  international conference on 3D vision (3DV). pp. 239--248. IEEE (2016)

\bibitem{lazaros2008review}
Lazaros, N., Sirakoulis, G.C., Gasteratos, A.: Review of stereo vision
  algorithms: from software to hardware. International Journal of
  Optomechatronics  \textbf{2}(4),  435--462 (2008)

\bibitem{lecun2015deep}
LeCun, Y., Bengio, Y., Hinton, G.: Deep learning. nature  \textbf{521}(7553),
  436--444 (2015)

\bibitem{Li17iccv}
Li, J., Klein, R., Yao, A.: A two-streamed network for estimating fine-scaled
  depth maps from single rgb images. In: Proceedings of the IEEE International
  Conference on Computer Vision. pp. 3372--3380 (2017)

\bibitem{li2018closedform_FastPhotoStyle}
Li, Y., Liu, M.Y., Li, X., Yang, M.H., Kautz, J.: A closed-form solution to
  photorealistic image stylization. In: Ferrari, V., Hebert, M., Sminchisescu,
  C., Weiss, Y. (eds.) Computer Vision -- ECCV 2018. pp. 468--483. Springer
  International Publishing, Cham (2018)

\bibitem{Liu15cvpr}
Liu, F., Shen, C., Lin, G.: Deep convolutional neural fields for depth
  estimation from a single image. In: Proceedings of the IEEE conference on
  computer vision and pattern recognition. pp. 5162--5170 (2015)

\bibitem{Ma16iros}
Ma, F., Carlone, L., Ayaz, U., Karaman, S.: Sparse sensing for
  resource-constrained depth reconstruction. In: 2016 IEEE/RSJ International
  Conference on Intelligent Robots and Systems (IROS). pp. 96--103. IEEE (2016)

\bibitem{Ma19icra}
Ma, F., Cavalheiro, G.V., Karaman, S.: Self-supervised sparse-to-dense:
  Self-supervised depth completion from lidar and monocular camera. In: 2019
  International Conference on Robotics and Automation (ICRA). pp. 3288--3295.
  IEEE (2019)

\bibitem{Ma18icra}
Ma, F., Karaman, S.: Sparse-to-dense: Depth prediction from sparse depth
  samples and a single image. In: 2018 IEEE International Conference on
  Robotics and Automation (ICRA). pp.~1--8. IEEE (2018)

\bibitem{mur2017orb}
Mur-Artal, R., Tard{\'o}s, J.D.: Orb-slam2: An open-source slam system for
  monocular, stereo, and rgb-d cameras. IEEE Transactions on Robotics
  \textbf{33}(5),  1255--1262 (2017)

\bibitem{Silberman:ECCV12}
Nathan~Silberman, Derek~Hoiem, P.K., Fergus, R.: Indoor segmentation and
  support inference from rgbd images. In: ECCV (2012)

\bibitem{ten2018using}
ten Pas, A., Platt, R.: Using geometry to detect grasp poses in 3d point
  clouds. In: Robotics Research, pp. 307--324. Springer (2018)

\bibitem{ronneberger2015u}
Ronneberger, O., Fischer, P., Brox, T.: U-net: Convolutional networks for
  biomedical image segmentation. In: International Conference on Medical image
  computing and computer-assisted intervention. pp. 234--241. Springer (2015)

\bibitem{UNet}
Ronneberger, O., Fischer, P., Brox, T.: U-net: Convolutional networks for
  biomedical image segmentation. In: Navab, N., Hornegger, J., Wells, W.M.,
  Frangi, A.F. (eds.) Medical Image Computing and Computer-Assisted
  Intervention -- MICCAI 2015. pp. 234--241. Springer International Publishing,
  Cham (2015)

\bibitem{ruder2017overview}
Ruder, S.: An overview of multi-task learning in deep neural networks (2017)

\bibitem{sajjan2019cleargrasp}
Sajjan, S.S., Moore, M., Pan, M., Nagaraja, G., Lee, J., Zeng, A., Song, S.:
  Cleargrasp: 3d shape estimation of transparent objects for manipulation
  (2019)

\bibitem{scharstein2002taxonomy}
Scharstein, D., Szeliski, R.: A taxonomy and evaluation of dense two-frame
  stereo correspondence algorithms. International journal of computer vision
  \textbf{47}(1-3),  7--42 (2002)

\bibitem{schoettler2019deep}
Schoettler, G., Nair, A., Luo, J., Bahl, S., Ojea, J.A., Solowjow, E., Levine,
  S.: Deep reinforcement learning for industrial insertion tasks with visual
  inputs and natural rewards. arXiv preprint arXiv:1906.05841  (2019)

\bibitem{simonyan2014very}
Simonyan, K., Zisserman, A.: Very deep convolutional networks for large-scale
  image recognition. arXiv preprint arXiv:1409.1556  (2014)

\bibitem{GoogLeNet}
{Szegedy}, C., {Wei Liu}, {Yangqing Jia}, {Sermanet}, P., {Reed}, S.,
  {Anguelov}, D., {Erhan}, D., {Vanhoucke}, V., {Rabinovich}, A.: Going deeper
  with convolutions. In: 2015 IEEE Conference on Computer Vision and Pattern
  Recognition (CVPR). pp.~1--9 (June 2015). \doi{10.1109/CVPR.2015.7298594}

\bibitem{tippetts2016review}
Tippetts, B., Lee, D.J., Lillywhite, K., Archibald, J.: Review of stereo vision
  algorithms and their suitability for resource-limited systems. Journal of
  Real-Time Image Processing  \textbf{11}(1),  5--25 (2016)

\bibitem{Uhrig17ic3dv}
Uhrig, J., Schneider, N., Schneider, L., Franke, U., Brox, T., Geiger, A.:
  Sparsity invariant cnns. In: International Conference on 3D Vision (3DV)
  (2017)

\bibitem{image_matting}
Wang, J., Cohen, M.: Image and video matting: A survey. Foundations and Trends
  in Computer Graphics and Vision  \textbf{3},  97--175 (01 2007).
  \doi{10.1561/0600000019}

\bibitem{wu2015active}
Wu, K., Ranasinghe, R., Dissanayake, G.: Active recognition and pose estimation
  of household objects in clutter. In: 2015 IEEE International Conference on
  Robotics and Automation (ICRA). pp. 4230--4237. IEEE (2015)

\bibitem{Yang19cvpr}
Yang, Y., Wong, A., Soatto, S.: Dense depth posterior (ddp) from single image
  and sparse range. In: Proceedings of the IEEE Conference on Computer Vision
  and Pattern Recognition. pp. 3353--3362 (2019)

\bibitem{zeng2019tossingbot}
Zeng, A., Song, S., Lee, J., Rodriguez, A., Funkhouser, T.: Tossingbot:
  Learning to throw arbitrary objects with residual physics (2019)

\bibitem{zeng20163dmatch}
Zeng, A., Song, S., Nie{\ss}ner, M., Fisher, M., Xiao, J., Funkhouser, T.:
  3dmatch: Learning local geometric descriptors from rgb-d reconstructions. In:
  CVPR (2017)

\bibitem{Zhang18cvpr}
Zhang, Y., Funkhouser, T.: Deep depth completion of a single rgb-d image. In:
  Proceedings of the IEEE Conference on Computer Vision and Pattern
  Recognition. pp. 175--185 (2018)

\bibitem{zhou2017unsupervised}
Zhou, T., Brown, M., Snavely, N., Lowe, D.G.: Unsupervised learning of depth
  and ego-motion from video. In: Proceedings of the IEEE Conference on Computer
  Vision and Pattern Recognition. pp. 1851--1858 (2017)

\bibitem{Zhou17cvpr}
Zhou, T., Brown, M., Snavely, N., Lowe, D.G.: Unsupervised learning of depth
  and ego-motion from video. In: Proceedings of the IEEE Conference on Computer
  Vision and Pattern Recognition. pp. 1851--1858 (2017)

\end{thebibliography}
\end{document}